\documentclass{article} 
\usepackage{Depends/iclr2027_conference,times}

\usepackage{amsmath,amsfonts,bm}

\def\eqref#1{equation~\ref{#1}}

\def\1{\bm{1}}

\DeclareMathAlphabet{\mathsfit}{\encodingdefault}{\sfdefault}{m}{sl}
\SetMathAlphabet{\mathsfit}{bold}{\encodingdefault}{\sfdefault}{bx}{n}

\usepackage{hyperref}
\usepackage{url}
\usepackage{graphicx}
\usepackage{wrapfig}
\usepackage{multirow}
\usepackage{booktabs}
\usepackage[nameinlink,capitalise]{cleveref}
\usepackage{amssymb}
\usepackage{pifont}
\usepackage{subcaption}
\usepackage{booktabs}
\usepackage{colortbl}
\usepackage{titletoc}
\usepackage{array}

\title{Reasoning Beyond Imitation: Observation Conditioned Distillation for Video LLMs}

\iclrfinalcopy

\author{Junzhou Chen, $\;\;$ Jindong Wang, $\;\;$ Gang Zhou  \\
William \& Mary\\
\texttt{\{jchen57,jdw,gzhou\}@wm.edu} \\
}

\begin{document}

\maketitle
\lhead{}
\vspace{-10pt}

\begin{abstract}

Reasoning distillation offers a promising approach to transfer temporal reasoning capabilities of powerful Video-LLMs to open-source student models.
However, teacher and student Video-LLMs may construct different temporal observations from the same video due to the adoption of distinct video sampling pipelines.
A reasoning trajectory valid under the teacher's observations may rely on evidence inaccessible to the student and thus cannot be directly executed based on the student's observations.
We formalize this as a conditional target mismatch, thereby revealing the Reasoning Acquisition–Execution Gap.
To bridge this gap, we propose an acquisition-to-execution framework that first distills teacher trajectories into a student reasoning prior and then further optimizes the execution capability of this prior across multiple observations.
We further introduce Execution-Aware Reference Regularization, which retains the existing prior when it is executable under current observations, while allowing a greater degree of adaptive adjustment when it is not.
Experiments across multiple Video-LLM backbones and six video reasoning benchmarks show that explicitly optimizing observation-conditioned execution consistently outperforms supervised distillation and standard RL-based optimization, with gains extending to held-out out-of-domain benchmarks.

\end{abstract}

\begin{figure*}[h]
    \centering
    
    \begin{subfigure}{\linewidth}
        \centering
        \includegraphics[width=0.9\linewidth]{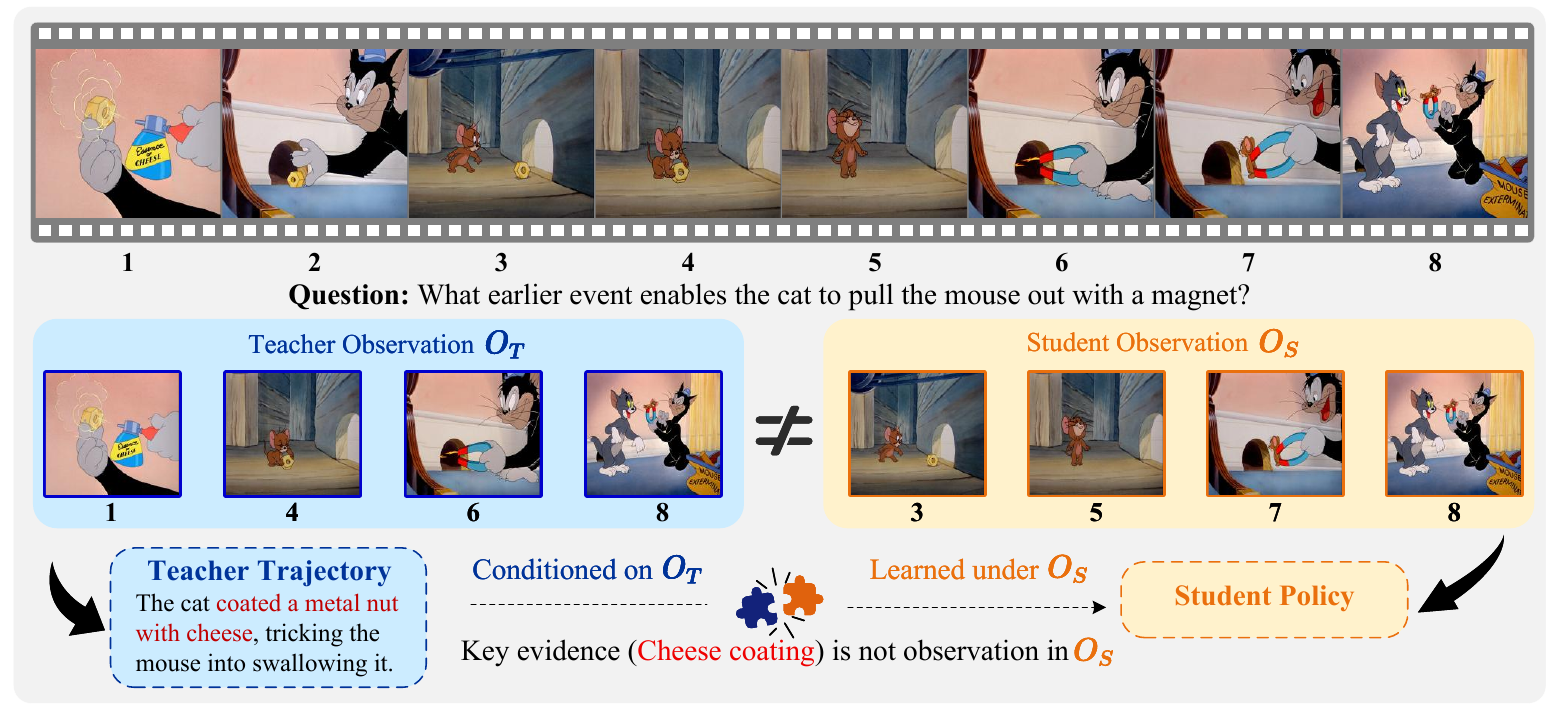}
        \caption{Teacher--student observation mismatch, making direct distillation not feasible.}
        \label{fig:sub_a}
    \end{subfigure}

    \begin{subfigure}{\linewidth}
        \centering
        \includegraphics[width=0.9\linewidth]{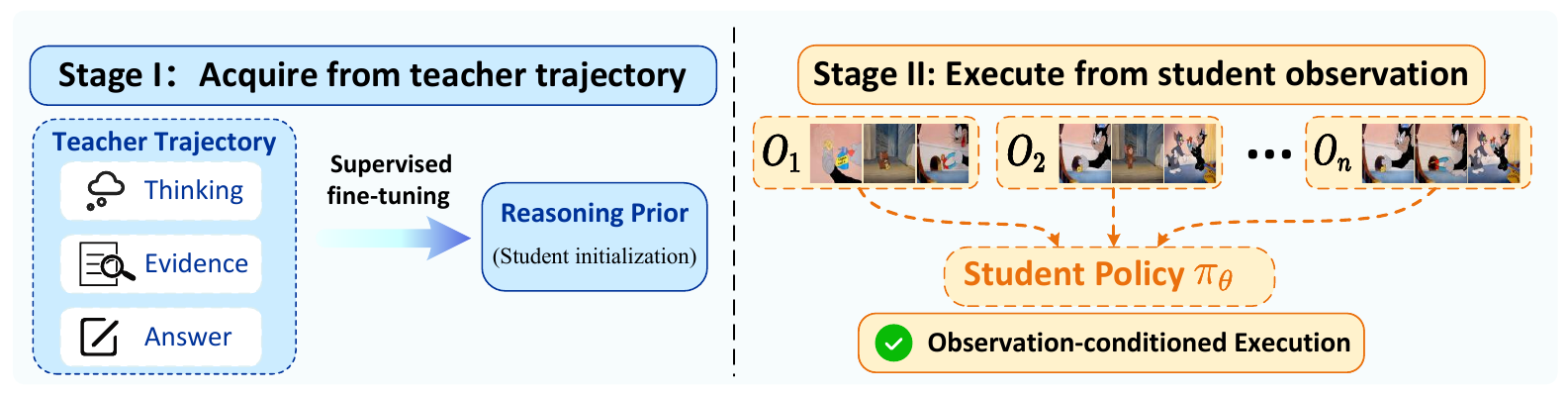}
        \caption{Overview of our cross-observation video reasoning distillation framework.}
        \label{fig:sub_b}
    \end{subfigure}
    
    \caption{Motivation and framework for observation-conditioned distillation for Video-LLMs.}
    \label{fig:intro}
\end{figure*}

\vspace{-20pt}
\section{Introduction}
\label{sec:introduction}
\vspace{-5pt}

Recently, proprietary Video Large Language Models (Video-LLMs)~\citep{anthropic2024claude3,comanici2025gemini,achiam2023gpt} have demonstrated significant advantages in complex video temporal understanding and reasoning tasks~\citep{cheng2025mcam, chen2025m, chen2026fast}, while open-source models still exhibit a substantial capability gap~\citep{li2026timeblind, krojer2025shortcut}. 
These strong teacher models are often accessible only through APIs, exposing only their outputs without providing model parameters, logits, or internal visual representations~\citep{jiang2023lion}. 
Therefore, black-box knowledge distillation~\citep{hinton2015distilling, guo2025deepseek, yan2026learning} based on generated reasoning trajectories emerges as a natural approach for capability transfer: reasoning trajectories from the teacher are used to supervise an open-source student, enabling it to acquire stronger video-reasoning capabilities. 
However, video scenarios introduce a challenge that differs fundamentally from text-based knowledge distillation: the teacher and the student may not be reasoning under the \textit{same} visual conditions.

As illustrated in \cref{fig:intro}, the same video does not necessarily imply the same model observation. Video-LLMs typically process only a limited sequence of frames obtained via temporal sampling, and different models often vary in their frame budgets, sampling rates, temporal offsets, and frame locations. 
As a result, even when given the same video and question, the teacher and student may reason over \textit{different temporal observations}. 
The key issue is that a teacher-generated trajectory is inherently conditioned on the teacher’s observation, whereas rationale distillation typically uses it directly as the supervision target under the student’s observation. 
Consequently, a reasoning path that is valid and effective given what the teacher observes may no longer remain valid or executable given what the student observes~\citep{wang2025internvl3, bai2025qwen3, achiam2023gpt, xu2025kdrl, li2023symbolic}. 
This implies that, in video reasoning distillation~\citep{agarwal2024policy, ko2024distillm}, the validity of the supervision target is itself observation-dependent, giving rise to a conditional target mismatch.

This observation-dependent mismatch further reveals a more fundamental limitation of rationale Supervised Fine-Tuning (SFT)~\citep{fu2026srft, ouyang2022training}. 
Although it can transfer teachers' reasoning trajectories to students, it does not directly address how to select and implement these behaviors based on the students' actual observations.
Consequently, acquiring teacher-demonstrated reasoning patterns does not necessarily translate into reliable observation-conditioned execution.
Such discrepancy can be referred to as the \textit{Reasoning Acquisition–Execution Gap}. 
This perspective reframes the problem from further improving teacher rationales to a different objective: transforming the distilled reasoning prior into reasoning capability that can be reliably executed from the student's own observations.

To address the above problem, we decompose video reasoning distillation into two stages: \textit{reasoning acquisition} and \textit{observation-conditioned} execution. 
In the first stage, the student is supervised with teacher-generated trajectories through SFT, yielding a reasoning prior. 
As shown in our ablation, this prior remains useful for subsequent execution optimization even when SFT alone does not improve task performance.
In Stage II, the student performs outcome-driven optimization over multiple temporal observations sampled from its own input stream, learning to execute reasoning behaviors conditioned on what it actually observes. 
Since the Stage-I prior can exhibit varying levels of executability across observations, we further introduce Execution-Aware Reference Regularization, which adaptively adjusts the reference constraint based on the prior’s executability in the current observation, 
balancing the preservation of already executable reasoning behaviors with the discovery of observation-specific alternative reasoning strategies.

Our main contributions are as follows:
\begin{itemize}
    \item We identify a conditional target mismatch induced by heterogeneous temporal observations and formulate the Reasoning Acquisition–Execution Gap to characterize the discrepancy between reasoning acquisition and observation-conditioned execution.
    \item We propose a two-stage acquisition-to-execution framework, where teacher-guided SFT is used to acquire a reasoning prior, followed by student-side Multi-Observation GRPO (MO-GRPO) to learn its observation-conditioned execution.
    \item We introduce Execution-Aware Reference Regularization, which adaptively adjusts the reference constraint based on the observation-wise executability of the reasoning prior, thereby balancing prior preservation and alternative reasoning discovery.
\end{itemize}

\section{Problem Formulation}
\label{sec:problem}

\begin{wrapfigure}{r}{0.65\textwidth}
\centering
\vspace{-.1in}
\includegraphics[width=\linewidth]{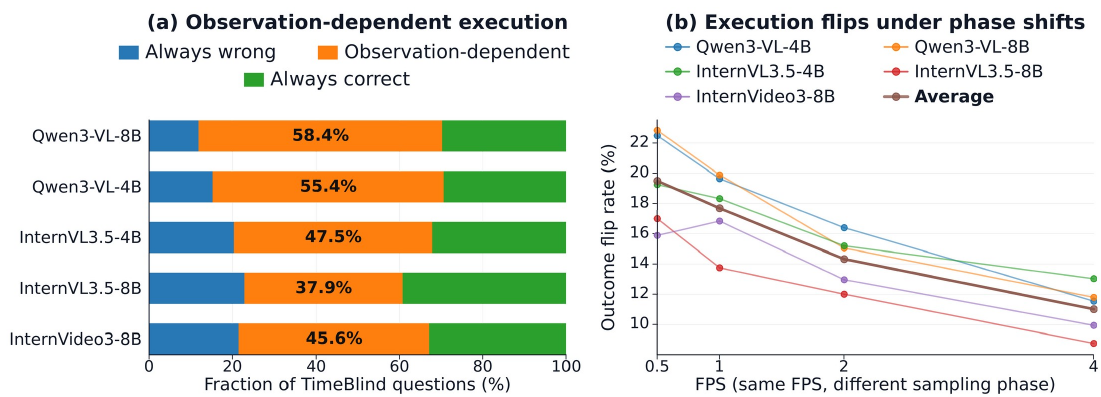}
\caption{
\textbf{Video-LLM execution varies across temporal observations on TimeBlind.} 
(a) Questions are categorized as always incorrect, observation-dependent, or always correct
across eight sampling configurations.
(b) Changing only the temporal sampling phase, while keeping the frame rate
fixed, still causes substantial outcome flips across models.
}
\vspace{-2mm}
\label{fig:observation_sensitivity}
\end{wrapfigure}

Given a video-question instance $(V,q,a^*)$, where $V$ is the raw video, $q$ the question, and $a^*$ the ground-truth answer, a Video-LLM reasons over a finite temporal observation rather than the complete video.
We denote the teacher and student observations as
$O_T=S_T(V;\xi_T)$ and $O_S=S_S(V;\xi_S)$,
where $\xi$ specifies the sampling configuration, such as frame budget, sampling rate, temporal offset, and frame locations.
Accordingly, the teacher and student may operate under different observation distributions, $p_T(O\mid V)$ and $p_S(O\mid V)$, and thus observe different evidence from the same video:
$O_T \neq O_S$.

Crucially, reasoning outcomes are not invariant to temporal observation.
Let $c_\pi(O,q)=\mathbb{I}[\operatorname{Eval}(\hat a_\pi(O,q),a^*)=1]$ indicate whether policy $\pi$ answers correctly under observation $O$.
If temporal sampling only changed input representation, we would expect
$c_\pi(O_1,q)=c_\pi(O_2,q)$ for different observations of the same video.
However, \cref{fig:observation_sensitivity} shows that many TimeBlind~\citep{li2026timeblind} questions are solved under only a subset of observations; the experimental details are provided in \cref{subsec:append_observation_sensitivity_details}.
Prediction flips remain frequent even when the frame rate is fixed and only the sampling phase changes, indicating that execution depends on \emph{which} temporal evidence is observed, not merely on the number of sampled frames.

This observation dependence creates a mismatch in reasoning distillation.
A teacher-generated trajectory is conditioned on the teacher observation,
$y_T\sim p_T(y\mid O_T,q)$,
whereas direct rationale distillation optimizes the student to reproduce that trajectory under a potentially different observation:
\begin{equation}
\mathcal{L}_{\mathrm{SFT}}=-\mathbb{E}_{(V,q,a^*)\sim\mathcal{D}}
\mathbb{E}_{O_T,O_S}
\left[
    \log \pi_\theta(y_T\mid O_S,q)
\right],
\qquad
y_T\sim p_T(\cdot\mid O_T,q).
\label{eq:mismatched_sft}
\end{equation}

The teacher trajectory may be fully valid under $O_T$ and still contain useful reasoning structure, yet its validity as a direct execution target is not guaranteed once the conditioning observation changes.
We refer to this discrepancy between the context under which supervision is generated and the context under which it is learned as \textbf{conditional target mismatch}.

At deployment, however, the student should maximize task performance over observations induced by its own sampler:
\begin{equation}
J_{\mathrm{exec}}(\pi)
=
\mathbb{E}_{(V,q,a^*)\sim\mathcal{D}}
\mathbb{E}_{O\sim p_S(O\mid V)}
\mathbb{E}_{y\sim\pi(\cdot\mid O,q)}
\left[
    R(y,a^*)
\right],
\label{eq:execution_objective}
\end{equation}
where $R(y,a^*)$ evaluates whether the generated trajectory leads to the correct task outcome.
Thus, rationale SFT transfers teacher-demonstrated reasoning behaviors, whereas successful execution requires selecting and realizing reasoning that is effective under the student's actual observation.
Minimizing imitation loss therefore does not necessarily optimize student-side execution.

We call this discrepancy the \textbf{Reasoning Acquisition--Execution Gap}.
Reasoning \emph{acquisition} learns reusable reasoning patterns from teacher demonstrations, while observation-conditioned \emph{execution} adapts those patterns to the evidence available in the current student observation.
This distinction is particularly important for video reasoning, where different temporal observations may expose different events and cues and therefore support different valid reasoning paths.

Accordingly, we denote the model obtained after rationale SFT as $\pi_{\mathrm{ref}}$ and treat it as a \textbf{reasoning prior}: it captures reasoning behaviors acquired from the teacher without assuming that they are optimally executable under every student observation.
Closing the acquisition--execution gap therefore requires further optimizing reasoning directly over student-side observations while preserving useful teacher-acquired structure.
This motivates our acquisition-to-execution framework in \cref{sec:method}.

\begin{figure*}[t!]
    \centering
    \includegraphics[width=\textwidth]{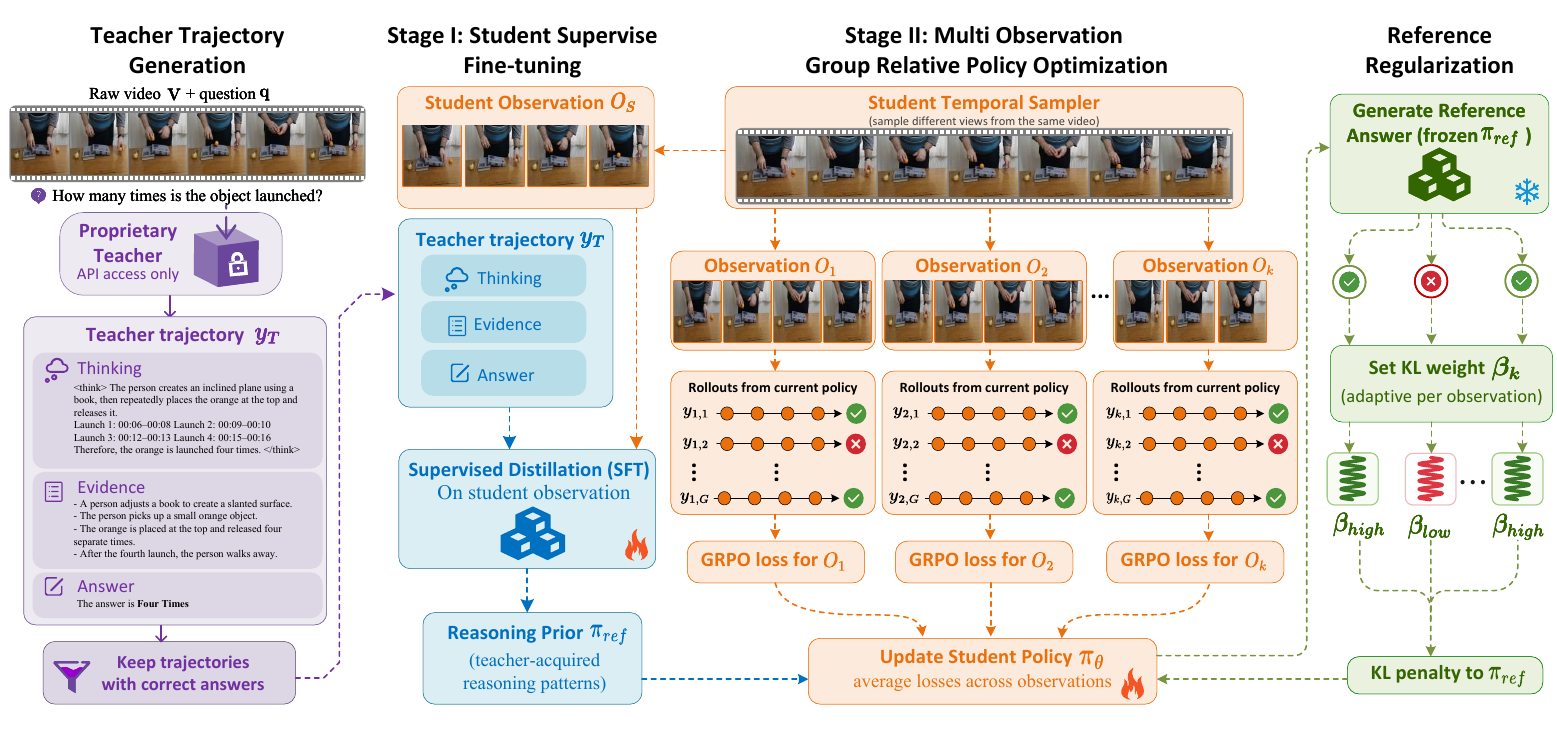}
    \caption{
    \textbf{Overview of our acquisition-to-execution framework.}
Stage I uses SFT on student observations to distill black-box teacher trajectories with correct answers into a reasoning prior.
Stage II computes GRPO losses separately for multiple temporal observations of the same video and aggregates them to update a shared student policy.
Reference regularization uses a larger KL weight when the frozen Stage-I model answers correctly under the current observation and a smaller weight otherwise.
}
    \label{fig:pipeline}
    \vspace{-2mm}
\end{figure*}

\section{Related Work}
\label{sec:related_work}

\paragraph{Video Reasoning in Video-LLMs}

Recently, the ability of Video-LLMs~\citep{team2025gemma, chen2024internvl,li2024llava} to perform complex video reasoning has been significantly enhanced, 
particularly in determining temporal sequences, analyzing event causality, tracking changes in object states, and understanding spatial relationships~\citep{kim2025videocomp, cheng2025mcam, kim2025videocomp, yuan2025videorefer}.
By breaking down video into timestamped frames and employing progressive multimodal training~\citep{an2026llava, ren2024timechat, feng2026video, wang2025internvl3, wang2025video}, the model is better equipped to handle video-related tasks involving localization and event sequencing.
As video complexity increases, Chain-of-Thought~\citep{fei2024video,zhang2025videollama, han2025videoespresso, chen2024advanced} breaks down complex video tasks into multiple perceptual and cognitive steps, enabling the model to arrive at the final answer progressively through an intermediate reasoning process.
Meanwhile, multiple video reasoning benchmarks~\cite{li2024mvbench, shangguan2025tomato, cheng2026v, qi2025vcr, upadhyay2026time, zhu2026mmr} indicate that high accuracy on traditional VideoQA tasks does not equate to reliable temporal reasoning capabilities. 
These studies demonstrate that, although current Video-LLMs possess strong video understanding capabilities, a significant gap in reasoning performance remains between proprietary and open-source models. 
Consequently, leveraging knowledge distillation to transfer the video reasoning capabilities of a powerful teacher model to a lightweight student model has emerged as an attractive research direction~\citep{guo2025deepseek, shi2025enhancing, lu2025vited, panigrahi2026good}.

\paragraph{Knowledge Distillation}

Knowledge distillation~\citep{gou2021knowledge, gu2024minillm, yang2025survey, zhang2026quest} primarily transfers knowledge from large models to smaller ones via prediction distributions or intermediate representations. 
Recently, Gradually extended the scope of distillation to the model's explicit reasoning process~\citep{li2023symbolic, lu2025vited}.
Reasoning distillation~\citep{hsieh2023distilling, wang2023scott, zhang2026quest, shridhar2023distilling, wang2026beyond} leverages rationales or CoT sequences generated by a teacher model to provide richer intermediate supervision, demonstrating the significant impact of explicit reasoning supervision on the performance of smaller models.
Offline teacher trajectories can lead to distribution discrepancies between the teacher and the student. 
Research~\citep{gu2024minillm, ko2024distillm, agarwal2024policy, ding2026videozoomer} has shifted the distillation process toward student-generated responses and integrated it with reinforcement learning. 
This concept has been extended to multimodal scenarios, incorporating on-policy supervision and reinforcement learning into multimodal knowledge distillation to enhance visual grounding and reasoning capabilities~\citep{xu2025kdrl, shi2025enhancing, luo2026agentark}.
However, we observe an \textit{observation mismatch} between teacher supervision and student learning. 
A correct reasoning trajectory from the teacher may not be directly grounded in the evidence visible to the student, resulting in suboptimal knowledge distillation performance.
\section{Method}
\label{sec:method}

To address the mentioned phenomenon described in~\cref{sec:problem}, we propose a video reasoning distillation framework as illustrated in~\cref{fig:pipeline}.
First, we set a proprietary video-LLM as a black-box teacher to generate video reasoning trajectories.
In stage I (\cref{subsec:stage_I}), the student model is supervised fine-turning on these trajectories to acquire the teacher's reasoning patterns.
Stage II (\cref{subsec:stage_II}), we construct multiple observation states and perform group relative policy optimization across these observations.
Moreover, since the Stage I prior can be executable under some observations but ineffective under others, we further introduce Execution-Aware Reference Regularization (\cref{subsec:OARR}), which adjusts the strength of the reference constraint separately for each observation.

\subsection{Stage I: Teacher Trajectory Distillation}
\label{subsec:stage_I}

\paragraph{Black-box teacher trajectory generation.}
Given a training sample $(V_i, q_i, a_i^*)$, where $V_i$ denotes the raw video, $q_i$ the question, and $a_i^*$ corresponding the ground truth answer.
The teacher obtains temporal observations using its own video sampler: $O_i^T = S_T(V_i, \xi_i^T)$. 
$\xi_i^T$ is the corresponding sampling configuration. 
Given $O_i^T$ and $q_i$, the teacher generates a structured reasoning trajectory 
\begin{equation}
    y_i^T = \left(T_i^T, E_i^T, \hat{a}_i^T\right)
\sim p_T\left(\cdot \mid O_i^T, q_i\right).
\end{equation}
$T_i^T$ denotes the thinking process generated by the teacher, $E_i^T$ represents the intermediate visual evidence referenced during reasoning, and $\hat{a}_i^T$ is the final predicted answer.
To avoid misleading students with erroneous trajectories, we retain only those trajectories that yield valid answers—specifically, those satisfying $\mathrm{Eval}\left(\hat{a}_i^T, a_i^*\right) = 1.$

\paragraph{Student Supervised Fine-tuning.}

Given the same video, the student uses its own video sampler to obtain observation $O_i^S = S_S(V_i, \xi_i^S)$.
Since the teacher and student may use different sampling strategies, $O_i^S$ is not equal to $O_i^T$.
We optimize the rationale-distillation objective in \cref{eq:mismatched_sft} over the retained correct teacher trajectories, instantiated with standard autoregressive token-level supervision.
Importantly, filtering by answer correctness only ensures that a teacher trajectory is valid under $O_i^T$; it does not guarantee that the same trajectory is directly executable from $O_i^S$.
Stage~I therefore deliberately uses the teacher trajectory as an \emph{acquisition signal}, despite the observation mismatch formalized in \cref{sec:problem}.
The objective of this stage is therefore not to establish $y_i^T$ as the unique reasoning path that should be reproduced under every student observation.
Instead, teacher trajectories expose the student to structured problem decomposition, temporal evidence organization, and multi-step inference patterns that may be reusable beyond the specific observation from which they were generated.
After Stage~I, we use the resulting model to initialize Stage~II and freeze an identical copy as the reference policy, denoted as $\pi_{\mathrm{ref}}$.



\subsection{Stage II: Multi-Observation Group Relative Policy Optimization}
\label{subsec:stage_II}

Stage II further optimizes the student under its own video sampling distribution. For each training example $(V_i,q_i,a_i^*)$, we construct $K$ temporal observations using the student sampler, denoted as $O_{i,k}=S_S(V_i;\xi_{i,k})$, where $k=1,\ldots,K$. 
Different $\xi_{i,k}$ correspond to different video sampling configurations, such as different sampling phases or frame locations. 

These observations correspond to the same video-question pair, but may expose different visual evidence to the model due to their different video sampling positions. 
For each $O_{i,k}$, the behavior policy $\pi_{\theta_{\mathrm{old}}}$ independently samples $G$ reasoning trajectories $y_{i,k,g}$. 
We do not require consistent reasoning trajectories across different observations, since different temporal views may support distinct yet equally valid reasoning paths.
For each rollout $y_{i,k,g}=(R_{i,k,g},E_{i,k,g},\hat{a}_{i,k,g})$, we compute the reward based on the final answer and output format:
\begin{equation}
    r_{i,k,g}=\lambda_{\mathrm{ans}} R_{\mathrm{ans}}(\hat{a}_{i,k,g},a_i^*)+\lambda_{\mathrm{fmt}} R_{\mathrm{fmt}}(y_{i,k,g}).
\end{equation}

$R_{\mathrm{ans}}$ provides the primary optimization signal based on the correctness of the final answer, while $R_{\mathrm{fmt}}$ maintains the structured reasoning--evidence--answer output format used in Stage I.

\paragraph{Observation-Local Relative Optimization.}
For each temporal observation $O_{i,k}$, we perform relative comparison only among the $G$ trajectories generated under the same observation. 
Specifically, we compute the reward mean and standard deviation within each group: $\mu_{i,k}=\frac{1}{G}\sum_{g=1}^{G} r_{i,k,g}$, then $\sigma_{i,k}=\sqrt{\frac{1}{G}\sum_{g=1}^{G}\left(r_{i,k,g}-\mu_{i,k}\right)^2}$,
and define the corresponding group-relative advantage as $A_{i,k,g}=\frac{r_{i,k,g}-\mu_{i,k}}{\sigma_{i,k}+\epsilon}.$
Unlike normalizing all rollouts from the same video-question pair together, we keep the rollout groups of different temporal observations independent. 
Since different observations may contain different task-relevant visual evidence, comparing rewards across observations would conflate the variation caused by the observation itself with the quality differences among reasoning trajectories. 
As a result, a successful trajectory from $O_{i,k'}$ does not reduce the relative advantage of trajectories generated from $O_{i,k}$. 
When all $G$ trajectories under the same observation receive identical rewards, we set the advantages of that group to zero.

We then define the token-level importance ratio as 
\begin{equation}
  \rho_{i,k,g,t}(\theta)=\frac{\pi_{\theta}\left(y_{i,k,g,t}\mid O_{i,k},q_i,y_{i,k,g,<t} \right)}{\pi_{\theta_{\mathrm{old}}}\left(y_{i,k,g,t}\mid O_{i,k},q_i,y_{i,k,g,<t}\right)}. 
\end{equation}

For each observation group, we independently apply the clipped Group Relative Policy Optimization (GRPO)~\citep{guo2025deepseek} objective:
\begin{equation}
\mathcal{L}^{\mathrm{GRPO}}_{i,k}=-\mathbb{E}_{g,t}\left[\min\left(\rho_{i,k,g,t}A_{i,k,g},\operatorname{clip}\left(\rho_{i,k,g,t},1-\epsilon_{\mathrm{clip}},1+\epsilon_{\mathrm{clip}}\right)A_{i,k,g}\right)\right].
\end{equation}

The GRPO losses from all $K$ temporal observations are jointly used to update the same student model. 
In this way, multiple video views of the same video-question pair can all provide training signals for policy optimization, while each reasoning trajectory is compared only against alternatives generated from the same visual observation.

\subsection{Execution-Aware Reference Regularization}
\label{subsec:OARR}

Stage II uses the Stage-I model as a frozen reference policy $\pi_\theta$ to prevent the student from drifting excessively from the reasoning behaviors learned during supervised fine-tuning. 
However, different temporal observations may expose different visual evidence, and the same reference policy can therefore perform differently across observations. 
It may already answer the question correctly under one observation while failing under another. 
Using the same KL coefficient for all observations is thus suboptimal. 
When the reference policy succeeds under the current observation, we would like to better preserve its behavior; 
when it fails, the constraint should be relaxed to allow the student to explore alternative reasoning trajectories.

For each temporal observation $O_{i,k}$, we first obtain a deterministic response from the frozen reference policy.
We then determine whether the reference policy correctly answers the question under this observation: 
$e_{i,k}=\mathbb{I}\left[\operatorname{Eval}\left(\hat{a}_{i,k}^{\mathrm{ref}},a_i^*\right)=1\right].$ 
Based on this result, we assign an observation-specific KL coefficient: $\beta_{i,k}=\beta_{\mathrm{low}}+e_{i,k}\left(\beta_{\mathrm{high}}-\beta_{\mathrm{low}}\right)$, where $0 \leq \beta_{\mathrm{low}} < \beta_{\mathrm{high}}.$
When $e_{i,k}=1$, we use the larger coefficient $\beta_{\mathrm{high}}$ to better preserve the reference behavior that already succeeds under $O_{i,k}$. 
When $e_{i,k}=0$, we instead use $\beta_{\mathrm{low}}$, allowing the policy to deviate further from the reference and explore alternative reasoning trajectories under the current observation.

Let $\mathcal{L}_{i,k}^{\mathrm{KL}}(\pi_\theta,\pi_{\mathrm{ref}})$ denote the reference regularization term for observation $O_{i,k}$. The complete Stage-II objective is
\begin{equation}
    \mathcal{L}_{\mathrm{StageII}}=\mathbb{E}_{i}\left[\frac{1}{K}\sum_{k=1}^{K}\left(\mathcal{L}_{i,k}^{\mathrm{GRPO}}+\beta_{i,k}\mathcal{L}_{i,k}^{\mathrm{KL}}\right)\right].
\end{equation}
Stage II performs GRPO optimization independently for each temporal observation, while adapting the reference constraint based on the reference policy's performance at that observation.

\begin{figure*}[t!]
    \centering
    \includegraphics[width=\textwidth]{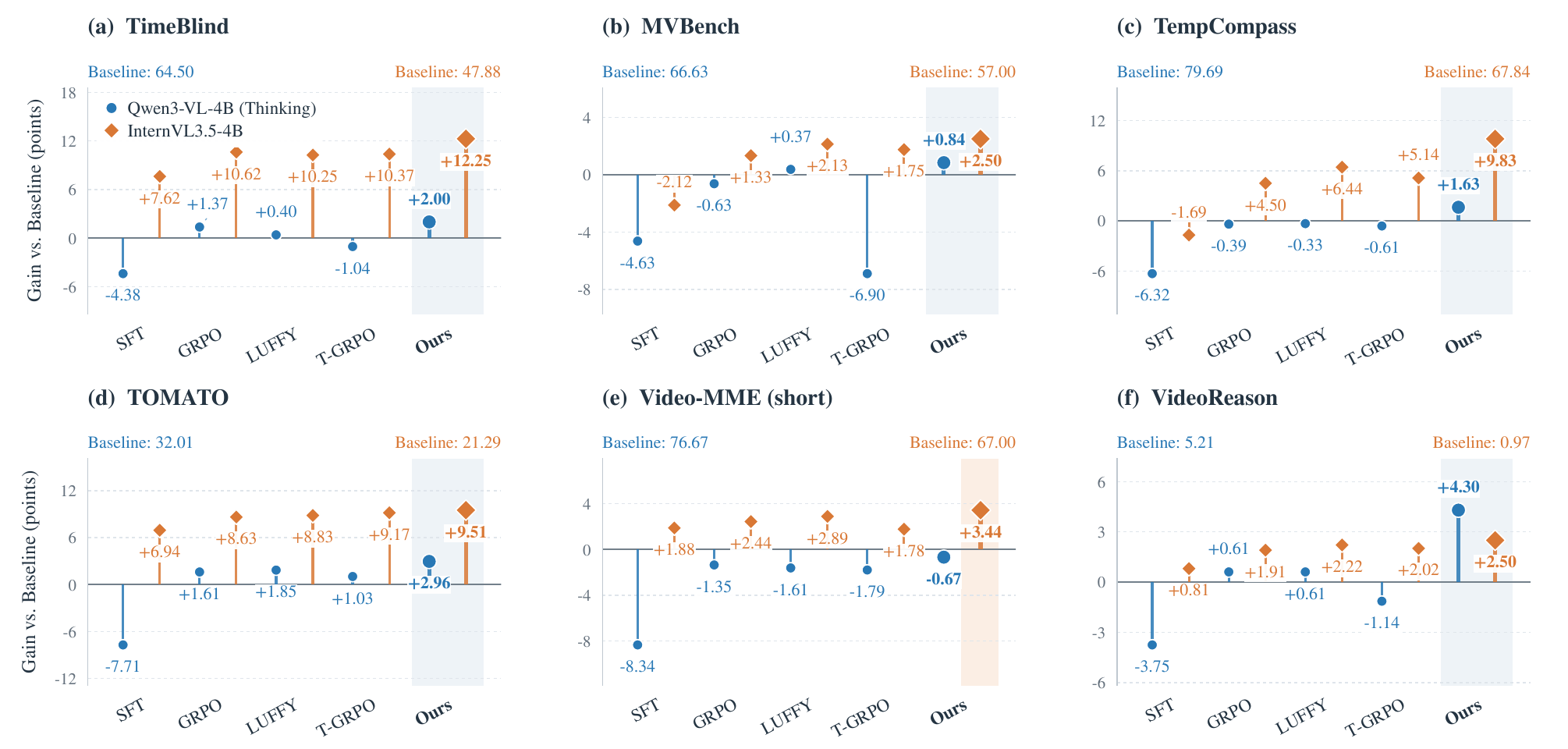}
    \caption{
    \textbf{Main comparison across two student reasoning regimes and six video reasoning benchmarks.}
    We compare SFT, GRPO, LUFFY, T-GRPO, and our method on Qwen3-VL-4B-Thinking and InternVL3.5-4B. 
    Each point shows the score change from the pretrained baseline, where positive values indicate improvement. Shaded regions denote the best method for each backbone. 
    Our method achieves the largest gain in 11 of 12 backbone-benchmark settings.
}
    \label{fig:compare}
    \vspace{-4mm}
\end{figure*}

\vspace{-2mm}

\section{Experiments}
\label{sec:experiments}

\vspace{-5pt}
\subsection{Experiment Setup}

\paragraph{Training and Evaluation Benchmarks.}
To comprehensively evaluate the effectiveness and generalization of our method, we conduct experiments on six video reasoning benchmarks, including three \textit{in-domain} benchmarks: 
TimeBlind~\citep{li2026timeblind}, MVBench~\citep{li2024mvbench}, and TempCompass~\citep{liu2024tempcompass}; 
and three \textit{out-of-domain} benchmarks: TOMATO~\citep{shangguan2025tomato}, Video-MME~\citep{fu2025video}, and VideoReasonBench~\citep{liu2026videoreasonbench}.
More details are provided in \cref{sec:append_data_generate}.

\vspace{-2mm}

\paragraph{Implementation Details.}

We use Gemini 2.5 Pro~\citep{comanici2025gemini} as the black-box teacher and Qwen3VL-{4B, 8B}~\citep{bai2025qwen3} and InternVL-3.5-{4B, 8B}~\citep{wang2025internvl3} as student models.
All students are fine-tuned with LoRA (\(r=32\), \(\alpha=64\)) in BF16.
Stage I and Stage II are trained for 2 and 1 epochs at learning rates of \(1\times10^{-5}\) and \(1\times10^{-6}\), respectively.
Full training hyperparameters and observation sampling configurations are in~\cref{sec:append_experiment}.
All RL-based methods are evaluated under the same compute budget.
We compare against standard SFT/GRPO pipelines and representative reasoning-distillation baselines; details are provided in~\cref{subsec:appendix_baselines}.

\begin{table*}[t!]
\centering
\caption{
\textbf{Scaling and generalization across student model families and sizes.} 
We report results for 4B and 8B variants of InternVL3.5 and Qwen3-VL-Thinking on three in-domain and three out-of-domain video reasoning benchmarks.
}
\vspace{-.1in}


\small
\setlength{\tabcolsep}{4.3pt}
\renewcommand{\arraystretch}{1.1}
\resizebox{\textwidth}{!}{
\begin{tabular}{cccccccc}
\toprule
                                  &                                       & \multicolumn{3}{c}{\textit{\textbf{In-Domain}}}                                                                          & \multicolumn{3}{c}{\textit{\textbf{Out-of-Domain}}}                                                                      \\ \cmidrule(lr){3-5}\cmidrule(lr){6-8}
\multirow{-2}{*}{\textbf{Params}} & \multirow{-2}{*}{\textbf{Method}}     & TimeBlind                              & MVBench                                & TempCompass                            & TOMATO                                 & Video-MME                              & VideoReason                            \\ \midrule
\multicolumn{8}{c}{\textit{\textbf{Proprietary Teacher Model: Gemini 2.5 Pro}}}                                                                                                                                                                                                                                                 \\ \hline
-                                 & -                                     & 81.13                                  & 72.25                                  & 83.85                                  & 46.91                                  & 90.56                                  & 56.00                                  \\ \hline
\multicolumn{8}{c}{\textit{\textbf{Student Model: InternVL 3.5}}}                                                                                                                                                                                                                                                               \\ \hline
                                  & Baseline                              & 47.88                                  & 57.00                                  & 67.84                                  & 21.29                                  & 67.00                                  & 0.97                                   \\
                                  & CoT SFT                               & 55.50                                  & 54.88                                  & 66.15                                  & 28.23                                  & 68.88                                  & 1.78                                   \\
                                  & SFT + GRPO                            & 58.50                                  & 58.33                                  & 72.34                                  & 29.92                                  & 69.44                                  & 2.88                                   \\
\multirow{-4}{*}{4B}              & \cellcolor[HTML]{EFEFEF}\textbf{Ours} & \cellcolor[HTML]{EFEFEF}\textbf{60.13} & \cellcolor[HTML]{EFEFEF}\textbf{59.50} & \cellcolor[HTML]{EFEFEF}\textbf{77.67} & \cellcolor[HTML]{EFEFEF}\textbf{30.80} & \cellcolor[HTML]{EFEFEF}\textbf{70.44} & \cellcolor[HTML]{EFEFEF}\textbf{3.47}  \\ \hline
                                  & Baseline                              & 64.38                                  & 62.50                                  & 77.47                                  & 26.62                                  & 73.43                                  & 1.32                                   \\
                                  & CoT SFT                               & 63.38                                  & 58.63                                  & 76.37                                  & 31.27                                  & 68.11                                  & 1.39                                   \\
                                  & SFT + GRPO                            & 65.25                                  & 60.62                                  & 76.95                                  & 33.62                                  & 72.44                                  & 2.43                                   \\
\multirow{-4}{*}{8B}              & \cellcolor[HTML]{EFEFEF}\textbf{Ours} & \cellcolor[HTML]{EFEFEF}\textbf{66.50} & \cellcolor[HTML]{EFEFEF}\textbf{62.88} & \cellcolor[HTML]{EFEFEF}\textbf{79.10} & \cellcolor[HTML]{EFEFEF}\textbf{34.23} & \cellcolor[HTML]{EFEFEF}\textbf{75.45} & \cellcolor[HTML]{EFEFEF}\textbf{4.12}  \\ \hline
\multicolumn{8}{c}{\textit{\textbf{Student Model: Qwen3-VL-Thinking}}}                                                                                                                                                                                                                                                          \\ \hline
                                  & Baseline                              & 64.50                                  & 66.63                                  & 79.69                                  & 32.01                                  & \textbf{76.67}                         & 5.21                                   \\
                                  & SFT                                   & 60.12                                  & 62.00                                  & 73.37                                  & 24.30                                  & 68.33                                  & 1.46                                   \\
                                  & SFT + GRPO                                  & 65.87                                  & 66.00                                  & 79.30                                  & 33.62                                  & 75.32                                  & 5.82                                   \\
\multirow{-4}{*}{4B}              & \cellcolor[HTML]{EFEFEF}\textbf{Ours} & \cellcolor[HTML]{EFEFEF}\textbf{66.50} & \cellcolor[HTML]{EFEFEF}\textbf{67.47} & \cellcolor[HTML]{EFEFEF}\textbf{81.32} & \cellcolor[HTML]{EFEFEF}\textbf{34.97} & \cellcolor[HTML]{EFEFEF}76.00          & \cellcolor[HTML]{EFEFEF}\textbf{9.51}  \\ \hline
                                  & Baseline                              & 67.00                                  & 65.38                                  & 80.34                                  & 34.57                                  & 78.56                                  & 6.11                                   \\
                                  & SFT                                   & 64.72                                  & 64.28                                  & 75.55                                  & 31.66                                  & 71.56                                  & 3.54                                   \\
                                  & SFT + GRPO                            & 67.88                                  & 67.75                                  & 79.86                                  & 35.37                                  & \textbf{81.04}                         & 7.36                                   \\
\multirow{-4}{*}{8B}              & \cellcolor[HTML]{EFEFEF}\textbf{Ours} & \cellcolor[HTML]{EFEFEF}\textbf{69.11} & \cellcolor[HTML]{EFEFEF}\textbf{68.62} & \cellcolor[HTML]{EFEFEF}\textbf{81.34} & \cellcolor[HTML]{EFEFEF}\textbf{36.74} & \cellcolor[HTML]{EFEFEF}80.22          & \cellcolor[HTML]{EFEFEF}\textbf{11.04} \\ \bottomrule
\end{tabular}}
\label{tab:main_result}

\vspace{-2mm}
\end{table*}

\subsection{Main Result}
\label{subsec:main_result}

\paragraph{Overall Comparison.}
\cref{fig:compare} compares our method with direct SFT, standard GRPO~\citep{guo2025deepseek}, LUFFY~\citep{yan2026learning}, and T-GRPO~\citep{feng2026video} on Qwen3-VL-4B-Thinking~\citep{bai2025qwen3} and InternVL3.5-4B~\citep{wang2025internvl3} across six video reasoning benchmarks. 
Our method achieves the strongest post-training performance across all twelve backbone–benchmark settings, and surpasses the corresponding pretrained baseline in eleven of them. 
In contrast, the competing methods show substantially less consistent behavior across backbones and tasks: improvements on one setting often do not transfer to another. 
This consistency is particularly notable given the two students' different reasoning behaviors, suggesting that the benefit of our method is not tied to a specific model family or reasoning initialization.
The same trend is reproduced with the open-source Qwen3-VL-235B-A22B-Thinking teacher as shown in \cref{subsec:opensource_teacher}.

\paragraph{Direct Trajectory Distillation.}
Direct teacher-trajectory SFT exhibits markedly different behavior across the two students. 
On InternVL3.5, SFT improves performance on several benchmarks, suggesting that teacher trajectories can provide useful structured reasoning supervision when the student does not rely on an explicit native thinking process. 
In contrast, SFT consistently degrades Qwen3-VL-Thinking across all six benchmarks, where the student already possesses its own explicit reasoning behavior. 
This contrast suggests that direct reasoning distillation is not uniformly beneficial: teacher trajectories may introduce useful reasoning structure for one student, while requiring another to reshape an already established reasoning policy.


\paragraph{RL-Based Baselines.}
RL-based optimization consistently improves upon direct trajectory SFT, with standard GRPO recovering much of the performance lost through direct imitation. 
More specialized approaches provide further gains in some settings: LUFFY~\citep{yan2026learning} incorporates off-policy teacher guidance, while T-GRPO~\citep{feng2026video} explicitly encourages sensitivity to temporal ordering. 
However, their improvements vary substantially across students and benchmarks, and neither consistently dominates standard GRPO. 
In contrast, our method remains the strongest post-training approach across all settings in ~\cref{fig:compare} by optimizing reasoning under multiple student-side temporal observations rather than treating execution as observation-invariant.

\vspace{-5pt}

\subsection{Scaling and Generalization}

\textbf{Scaling Across Model Sizes.}
\cref{tab:main_result} extends the main comparison to both 4B and 8B students. The gains from observation-conditioned execution persist as the student scales up: our method improves over the corresponding SFT+GRPO pipeline in all 4B settings and in all but one 8B setting. 
The improvement therefore does not disappear as the underlying student becomes stronger. 
Notably, post-training also closes a substantial part of the capacity gap: on several benchmarks, the 4B model trained with our method matches or exceeds the corresponding pretrained 8B model, showing that stronger reasoning execution can compensate for part of the difference in model scale.

\textbf{Out-of-Domain Generalization.}
The improvement also transfers beyond the benchmarks used for training. While Stage I and Stage II are trained on TimeBlind~\citep{li2026timeblind}, MVBench~\citep{li2024mvbench}, and TempCompass~\citep{liu2024tempcompass}, the same advantage is retained on TOMATO~\citep{shangguan2025tomato}, Video-MME (Short)~\citep{fu2025video}, and VideoReasonBench~\citep{liu2026videoreasonbench}, which are held out entirely from training. Across the four student backbones, our method improves over standard SFT+GRPO in all in-domain settings and in all but one out-of-domain setting. This consistency across benchmarks with different task formats and reasoning requirements suggests that the learned execution behavior is not restricted to the training tasks.

\subsection{Component Ablation}
\label{subsec:ablation_study}

\begin{wrapfigure}{r}{0.6\textwidth}
\centering

\vspace{-10pt}

\includegraphics[width=\linewidth]{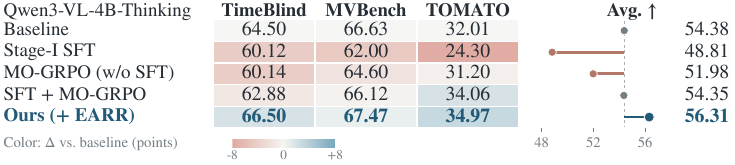}
\caption{\textbf{Ablation study of the proposed framework.} We evaluate the contributions of Stage-I SFT, multi-observation GRPO, and execution-aware reference regularization.
}

\vspace{-10pt}

\label{fig:ablation}
\end{wrapfigure}

\paragraph{Complementary Acquisition and Execution.}
\cref{fig:ablation} isolates the roles of teacher-guided reasoning acquisition and student-side execution optimization. 
Neither Stage-I SFT nor multi-observation GRPO is sufficient on its own: direct trajectory distillation degrades the pretrained model, while applying multi-observation optimization without the teacher-acquired prior only partially recovers its performance. 
Combining the two substantially improves all evaluated benchmarks, indicating that the two stages serve complementary roles rather than interchangeable ones. 
Stage I provides structured reasoning patterns acquired from the teacher, whereas Stage II adapts these patterns to reasoning trajectories that are effective under the student’s own temporal observations.

\vspace{-5pt}

\paragraph{Execution-Aware Regularization.}
Adding execution-aware reference regularization further improves the combined model across all evaluated benchmarks. This result shows that preserving the Stage-I reasoning prior with a uniform constraint is still suboptimal: the prior may already be effective under one temporal observation but fail under another. EARR explicitly accounts for this observation-wise difference, preserving successful reasoning behavior while allowing larger policy updates when the prior is not executable. The improvement is most pronounced on TimeBlind, which is also the benchmark where ~\cref{fig:observation_sensitivity} shows substantial sensitivity to temporal sampling, making the result consistent with the motivation for observation-dependent reference control.

\subsection{Execution Across Temporal Observations}


\begin{wraptable}{r}{0.5\textwidth}
\vspace{-1.0em}
\centering
\caption{\textbf{Multi-observation ablation.}
Performance under different temporal observations in the TimeBlind benchmark, where $\phi$ denotes the temporal sampling phase.}
\label{tab:ablation_observation}
\vspace{-0.4em}

\footnotesize
\setlength{\tabcolsep}{4.2pt}
\renewcommand{\arraystretch}{1.05}
\begin{tabular}{lccccc}
\hline
                         & \multicolumn{2}{c}{\textbf{FPS = 1}} & \multicolumn{3}{c}{\textbf{FPS = 2}}             \\ \cmidrule(lr){2-3} \cmidrule(lr){4-6}
\multirow{-2}{*}{Method} & $\phi{=}0$        & $\phi{=}.5$      & $\phi{=}0$     & $\phi{=}.25$   & $\phi{=}.5$    \\ \hline
Baseline                 & 56.75             & 60.75            & 64.50          & 64.25          & 63.88          \\
SFT                      & 53.75             & 52.50            & 60.12          & 58.62          & 59.00          \\
GRPO                     & 54.50             & 56.12            & 65.88          & 64.50          & 65.88          \\
\rowcolor[HTML]{EFEFEF} 
\textbf{Ours}            & \textbf{62.75}    & \textbf{63.38}   & \textbf{66.50} & \textbf{67.88} & \textbf{67.75} \\ \hline
\end{tabular}

\vspace{-0.6em}
\end{wraptable}

\cref{sec:problem} shows that reasoning execution can change with temporal observation, making a teacher trajectory useful for acquisition but not necessarily executable from every student-side view. 
We therefore evaluate Qwen3-VL-4B-Thinking~\citep{bai2025qwen3} on TimeBlind~\citep{li2026timeblind} under the five observations used in Stage II. As shown in \cref{tab:ablation_observation}, direct SFT remains sensitive to temporal sampling. 
Standard GRPO partially recovers this loss, but optimization under a single observation does not consistently transfer when the observed frames change. 
In contrast, our method improves execution across different observations by explicitly optimizing the student on multiple views and conditioning the reference constraint on whether the acquired prior is executable under each view. 
The case study in \cref{sec:append_case_study} further shows that this does not require reproducing the same reasoning trajectory: 
our model uses different observation-specific evidence while reaching the correct conclusion.

\section{Conclusion}



In this paper, we identify the Reasoning Acquisition--Execution Gap in video reasoning distillation, where teacher-generated reasoning may not be directly executable under the student's temporal observations. 
To address this issue, we propose an acquisition-to-execution framework that first distills teacher reasoning into a student prior and then optimizes its execution across multiple student-side observations with execution-aware reference regularization. 
Extensive experiments across multiple Video-LLM backbones and reasoning benchmarks demonstrate consistent improvements over supervised distillation and standard RL-based optimization. 
These results highlight that effective video reasoning distillation requires not only acquiring teacher reasoning, but also adapting its execution to the evidence actually observed by the student.
We study reasoning distillation for Video-LLMs under heterogeneous temporal observations.

\clearpage

\subsection*{AI use statement}

We used generative AI tools to assist with manuscript revision, literature search, and code development.
The authors take responsibility for the accuracy and integrity of the final manuscript.

\subsection*{Reproducibility statement}

We provide detailed information to support reproducibility throughout the appendix. 
\cref{sec:append_data_generate} describes the dataset construction, training/evaluation splits, and teacher trajectory generation procedure; 
\cref{sec:append_experiment} provides the model configurations, training hyperparameters, temporal sampling settings, evaluation protocols, and prompts. 
Additional analyses and experiments, including reproduction with an open-source teacher model, are provided in \cref{sec:append_experiment}.



\bibliography{References}
\bibliographystyle{Depends/iclr2027_conference}




\clearpage
\startcontents[appendix]

\section*{Appendix Contents}
\printcontents[appendix]{}{1}{}

\clearpage


\appendix
\section{Outline of the appendix}
Below, we outline the structure of the appendix.

\cref{sec:append_related_work} provides additional discussion of related work on video reasoning and knowledge distillation.
\cref{sec:append_data_generate} details the data construction pipeline, including dataset composition, teacher trajectory generation, and the construction of the Stage-II RL training set.
\cref{sec:append_experiment} provides complete experimental details, including model configurations, training hyperparameters, baselines, evaluation protocols, and prompts.
Finally, \cref{sec:append_case_study} presents qualitative case studies on reasoning under different temporal observations and practical observation mismatch across Video-LLMs.

\section{Full Related Work}
\label{sec:append_related_work}

\subsection{Video Reasoning in Video-LLMs}

Recently, the ability of Video-LLMs~\citep{team2025gemma, chen2024internvl,li2024llava} to perform complex video reasoning has been significantly enhanced, 
particularly in determining temporal sequences, analyzing event causality, tracking changes in object states, and understanding spatial relationships~\citep{kim2025videocomp, cheng2025mcam, kim2025videocomp, yuan2025videorefer}.
By breaking down video into timestamped frames and employing progressive multimodal training~\citep{an2026llava, ren2024timechat, feng2026video, wang2025internvl3, wang2025video}, the model is better equipped to handle video-related tasks involving localization and event sequencing.

As video complexity increases, Chain-of-Thought~\citep{fei2024video,zhang2025videollama, han2025videoespresso, chen2024advanced} breaks down complex video tasks into multiple perceptual and cognitive steps, enabling the model to arrive at the final answer progressively through an intermediate reasoning process.
Meanwhile, multiple video reasoning benchmarks~\citep{li2024mvbench, shangguan2025tomato, cheng2026v, qi2025vcr, upadhyay2026time, zhu2026mmr} indicate that high accuracy on traditional VideoQA tasks does not equate to reliable temporal reasoning capabilities. 
These studies demonstrate that, although current Video-LLMs possess strong video understanding capabilities, a significant gap in reasoning performance remains between proprietary and open-source models. 

Many studies~\citep{shi2025enhancing, lu2025vited, wang2024dancing, you2026adaptive} focus directly on optimizing the reasoning capabilities of video LLMs.
Video-R1~\citep{feng2026video} introduces reinforcement learning to video reasoning, employing a temporally aware reward design to encourage the model to more fully utilize temporal information from the video during the reasoning process.
Thinking with Videos~\citep{zhang2026thinking} enables models to actively locate and resample relevant video segments during inference.
VideoP2R~\citep{jiang2026videop2r} explicitly distinguishes between perception and reasoning processes and introduces more fine-grained process supervision during reinforcement learning.
These methods primarily investigate how a single Video-LLM can more effectively use the video information it has acquired to reason. 
In contrast, we focus on a distinct issue arising in knowledge transfer for cross-model reasoning: the teacher and student may obtain different temporal observations due to differing video sampling processes; 
consequently, a reasoning process valid for the teacher cannot necessarily be directly executed based on the video evidence actually seen by the student.

\subsection{Knowledge Distillation}

Knowledge distillation~\citep{gou2021knowledge, gu2024minillm, yang2025survey, zhang2026quest, hinton2015distilling} primarily transfers knowledge from large models to smaller ones via prediction distributions or intermediate representations. 
Recently, the scope of distillation has gradually been extended to the model's explicit reasoning process~\citep{li2023symbolic, lu2025vited}.
Reasoning distillation~\citep{hsieh2023distilling, wang2023scott, zhang2026quest, shridhar2023distilling, wang2026beyond} leverages rationales or CoT sequences generated by a teacher model to provide richer intermediate supervision, demonstrating the significant impact of explicit reasoning supervision on the performance of smaller models.
Offline teacher trajectories can lead to distribution discrepancies between the teacher and the student. 

On the other hand, relying on fixed reasoning trajectories generated offline by the teacher introduces new challenges~\citep{zhang2026quest, shridhar2023distilling}. 
During training, the student is required to fit teacher-generated sequences, whereas at inference time it generates outputs under its own policy, potentially leading to a substantial policy mismatch between the teacher and the student. 
MiniLLM~\citep{gu2024minillm} improves generative distillation for autoregressive language models through reverse KL optimization.

Knowledge distillation has also been extended to multimodal and video reasoning scenarios. 
Agent-of-Thoughts Distillation (AoTD)~\citep{shi2025enhancing} first employs multiple video agents with specialized capabilities to construct verified spatio-temporal reasoning chains, and then distills these reasoning capabilities into a single Video-LLM. 
VOLD~\citep{bousselham2026vold} combines GRPO with on-policy distillation to continuously incorporate teacher supervision during reinforcement learning for vision-language models, and highlights the importance of a reasoning-SFT cold start before online knowledge transfer.
Video-OPD~\citep{li2026video} leverages a stronger teacher to provide fine-grained token-level supervision on trajectories generated by the student itself.
These studies suggest that the conditions under which reasoning supervision is generated, and the conditions under which the student learns from that supervision, are becoming increasingly important in video reasoning distillation. 

Research~\citep{gu2024minillm, ko2024distillm, agarwal2024policy, ding2026videozoomer} has shifted the distillation process toward student-generated responses and integrated it with reinforcement learning. 
This concept has been extended to multimodal scenarios, incorporating on-policy supervision and reinforcement learning into multimodal knowledge distillation to enhance visual grounding and reasoning capabilities~\citep{xu2025kdrl, shi2025enhancing, luo2026agentark}.
Unlike existing approaches that distill from privileged visual evidence, we focus on the observation mismatch that naturally arises from differing temporal sampling strategies between the teacher and the student, where a reasoning trajectory valid under the teacher's observations may no longer be executable given the evidence actually observed by the student.


\begin{table*}[ht!]
\centering
\caption{\textbf{Representative temporal sampling pipelines in Video-LLMs.}
Different models construct temporal observations using different sampling
rules and frame constraints.}
\label{tab:video_sampling}
\scriptsize
\setlength{\tabcolsep}{7pt}
\renewcommand{\arraystretch}{1.15}

\begin{tabular}{llll}
\hline
\textbf{Model} & \textbf{Reference Setting} & \textbf{Frame Selection} & \textbf{Budget / Control} \\ \hline
Gemini 2.5 Pro & 1 FPS & Fixed-rate sampling & FPS configurable \\
Qwen3-VL & 2 FPS & FPS to frame count, then uniform indices & 4--768 frames or explicit frame count \\
InternVL3.5 & Fixed number of segments & One frame per temporal segment & Segment count configurable \\
LLaVA-Video & 1 FPS & FPS sampling or uniform sampling & Maximum frame count configurable \\
VideoLLaMA3 & 1 FPS & FPS sampling with uniform subsampling & Maximum frame count configurable \\ \hline
\end{tabular}

\end{table*}

\subsection{Temporal Observation Construction in Video-LLMs}
\label{sec:temporal_sampling}

Video-LLMs~\citep{chen2025live, zhang2025videollama, zhang2024llava, zhang2025videollama} do not share a standardized procedure for converting a raw video into the finite sequence of frames used for inference.
\cref{tab:video_sampling} summarizes representative temporal preprocessing pipelines documented in official model interfaces, model cards, and released code.
We emphasize that these configurations are not immutable properties of the models: several ystems allow users to modify the sampling rate or frame budget.
Rather, the comparison illustrates that the effective temporal observation is determined by a model-specific sampling pipeline, instead of a single universal FPS parameter.

As shown in \cref{tab:video_sampling}, these pipelines differ in both their parameterization and frame-selection rules.
Gemini's~\citep{comanici2025gemini} static video interface uses fixed-rate sampling by default,
whereas Qwen3-VL~\citep{bai2025qwen3} converts a requested FPS into a target number of frames and distributes the selected indices over the temporal range.
InternVL3.5~\citep{wang2025internvl3} instead exposes a segment-based video loader that divides the video into a configurable number of temporal intervals and samples one representative frame from each interval.
Other open-source Video-LLMs~\citep{zhang2024llava, zhang2025videollama} combine rate-based sampling with explicit frame budgets, switching or resampling to a fixed number of frames
when the budget becomes active.
Thus, two systems receiving the same raw video can construct different frame sequences even when their nominal sampling rates appear similar.

This distinction motivates treating the temporal observation as $O=S(V;\xi)$, where $\xi$ describes the complete sampling configuration rather than only the nominal frame rate.
In practice, $\xi$ may include the sampling rate, frame budget, temporal phase or starting offset, frame-index construction, and budget-induced resampling rule.
Consequently, matching the nominal FPS alone does not in general guarantee identical temporal observations across Video-LLMs.

\section{Data Generation Methods and Pipeline}
\label{sec:append_data_generate}

\subsection{Dataset Composition and Evaluation Split}

\begin{table*}[h!]
\centering
\caption{\textbf{Benchmark card.}
Summary of the video reasoning benchmarks used in our experiments.
``Scale'' reports the number of videos and question--answer pairs.
}
\vspace{2mm}
\label{tab:benchmark_card}

\small
\setlength{\tabcolsep}{5pt}
\renewcommand{\arraystretch}{1.15}
\resizebox{\textwidth}{!}{
\begin{tabular}{lllll}
\hline
\textbf{Benchmark}                                                                             & \textbf{Scale}                 & \textbf{Video Duration} & \textbf{Main Focus}                                                                                                                                                            & \textbf{Model Output}                                                                     \\ \hline
\textbf{\begin{tabular}[c]{@{}l@{}}TimeBlind\\ \citep{li2026timeblind}\end{tabular}}        & 600 instances / 2.4K QA        & 8.5s avg.               & \begin{tabular}[c]{@{}l@{}}Temporal Compositionality; \\ Event Order; Structural Logic;\\ Event Attribute;\end{tabular}                                                        & Two-choice answer                                                                         \\ \hline
\textbf{\begin{tabular}[c]{@{}l@{}}Video-MME\\ \citep{fu2025video}\end{tabular}}        & 900 videos / 2.7K QA           & 11s--60min              & \begin{tabular}[c]{@{}l@{}}Temporal Reasoning; \\ Action Reasoning; \\ Perception; Recognition; \\ Long-Video Understanding\end{tabular}                                       & Four-choice answer                                                                        \\ \hline
\textbf{\begin{tabular}[c]{@{}l@{}}TOMATO\\ \citep{shangguan2025tomato}\end{tabular}}           & 1.4K videos / 1.5K QA          & 9.2s avg.; 72.7s max.   & \begin{tabular}[c]{@{}l@{}}Action Count; Direction; Rotation; \\ Shape \& Trend; Visual Cues;\\ Velocity \& Frequency;\end{tabular}                                            & Multiple-choice answer                                                                    \\ \hline
\textbf{\begin{tabular}[c]{@{}l@{}}VideoReasonBench\\ \citep{liu2026videoreasonbench}\end{tabular}} & 240 videos / 1.4K QA           & 54.3s avg.; 154.8s max. & \begin{tabular}[c]{@{}l@{}}Operation Recall; State Tracking; \\ State Comparison; State Prediction; \\ Operation Planning\end{tabular}                                         & \begin{tabular}[c]{@{}l@{}}Open-ended structured state;\\ Operation sequence\end{tabular} \\ \hline
\textbf{\begin{tabular}[c]{@{}l@{}}MVBench\\ \citep{li2024mvbench}\end{tabular}}          & 3.6K Video clips / 4.0K QA     & $\sim$16s avg.          & \begin{tabular}[c]{@{}l@{}}Action Understanding; \\ Motion Understanding; \\ Object Interaction; State Change; \\ Episodic Reasoning; \\ Counterfactual Reasoning\end{tabular} & Multiple-choice answer                                                                    \\ \hline
\textbf{\begin{tabular}[c]{@{}l@{}}TempCompass\\ \citep{liu2024tempcompass}\end{tabular}}      & 410 videos / 7.5K instructions & $\sim$11.4s avg.        & \begin{tabular}[c]{@{}l@{}}Action; Direction; Speed; \\ Event Order; Attribute Change\end{tabular}                                                                             & Multiple-choice, yes/no,                                                                  \\ \hline
\end{tabular}
}
\end{table*}

We construct the training pool from three video reasoning benchmarks, as summarized in~\cref{tab:benchmark_card}:
TimeBlind~\citep{li2026timeblind},
MVBench~\citep{li2024mvbench},
and TempCompass~\citep{liu2024tempcompass}.
Specifically, we sample 1,600 video-question pairs from TimeBlind, 3,000 from MVBench,
and 4,000 from TempCompass.
For TempCompass, we exclude caption-oriented questions and retain only questions with deterministic answers.
This results in a total of 8,600 video-question pairs for teacher trajectory generation and subsequent training.

The remaining samples from these three benchmarks are reserved for in-domain evaluation and are never used during training.
We further ensure that the training and evaluation splits are \textbf{completely disjoint at the video level}, i.e., no video appears in both training and evaluation sets.
For out-of-domain evaluation, we use TOMATO~\citep{shangguan2025tomato},
the Short subset of Video-MME~\citep{fu2025video},
and VideoReasonBench~\citep{liu2026videoreasonbench}.
No samples from these benchmarks are used for either Stage-I supervised distillation or Stage-II reinforcement learning.
The dataset composition and training/evaluation usage are summarized in~\cref{tab:benchmark_split}.

\begin{table}[ht!]
\centering
\caption{\textbf{Dataset split and training usage.}
The training pool is constructed from TimeBlind, MVBench, and TempCompass.
Held-out samples from these benchmarks are used for in-domain evaluation,
while TOMATO, Video-MME (Short), and VideoReasonBench are used exclusively
for out-of-domain evaluation.}
\label{tab:benchmark_split}

\small
\setlength{\tabcolsep}{5pt}
\renewcommand{\arraystretch}{1.1}
\begin{tabular}{lcccc}
\hline
\textbf{Benchmark}
& \textbf{Available QA}
& \textbf{Training QA}
& \textbf{Training Exposure}
& \textbf{Domain} \\ \hline

TimeBlind
& 2,400
& \textbf{1,600}
& Yes
& In-Domain \\

MVBench
& 4,000
& \textbf{3,000}
& Yes
& In-Domain \\

TempCompass
& 5,500$^{*}$
& \textbf{4,000}
& Yes
& In-Domain \\ \hline

Video-MME (Short)
& 900
& 0
& No
& Out-of-Domain \\

TOMATO
& 1,500
& 0
& No
& Out-of-Domain \\

VideoReasonBench
& 1,400
& 0
& No
& Out-of-Domain \\ \hline

\textbf{Total}
& 15,600
& \textbf{8,600}
& --
& -- \\ \hline
\end{tabular}

\vspace{2pt}
\begin{minipage}{0.96\linewidth}
\footnotesize
$^{*}$Caption-related questions are excluded from TempCompass, leaving approximately 5.5K objective QA samples.

$^{\dagger}$Only the Short subset of Video-MME is used for evaluation.
$N_{\mathrm{short}}$ denotes the number of QA pairs in the evaluated Short subset.
\end{minipage}

\end{table}

\subsection{Multi-Configuration Teacher Trajectory Generation}

For each training video-question pair, we generate four reasoning trajectories using the black-box teacher model.
The four generations use different generation configurations to expose the student to diverse teacher reasoning behaviors for the same underlying problem.
Across configurations, we keep the task prompt and required output format fixed, with only minimal benchmark-specific changes such as formatting multiple-choice options.
The complete teacher response, including the reasoning trajectory, evidence, final answer, and generation metadata, is stored for every rollout.

Importantly, we distinguish between \emph{raw trajectory generation} and the trajectories used for Stage-I supervised distillation.
All four raw teacher generations are retained so that their correctness can later be used to construct the Stage-II training set.
For Stage-I SFT, however, we only use teacher trajectories whose final answers are correct according to the benchmark ground truth.
Incorrect teacher generations are therefore never used as supervised reasoning targets, but their correctness labels are retained for identifying examples on which the teacher reasoning process is not consistently successful.

Formally, let
\[
\mathcal{Y}_i^{T}=\{y_{i,1}^{T},\ldots,y_{i,4}^{T}\}
\]
denote the four teacher trajectories generated for training example $x_i=(V_i,q_i,a_i^*)$,
and let $\hat{a}_{i,j}^{T}$ denote the final answer of the $j$-th trajectory.
The set of valid Stage-I supervision trajectories is
\begin{equation}
\mathcal{Y}_{i,\mathrm{SFT}}^{T}
=
\left\{
y_{i,j}^{T}
\in
\mathcal{Y}_i^{T}
\;\middle|\;
\operatorname{Eval}\!\left(\hat{a}_{i,j}^{T},a_i^*\right)=1
\right\}.
\end{equation}
Only trajectories in $\mathcal{Y}_{i,\mathrm{SFT}}^{T}$ are used as supervised targets during Stage I.

\subsection{Error-Focused RL Dataset Construction}
\label{subsec:append_rl_dataset}
Stage II focuses reinforcement learning on training examples for which the teacher generations are not uniformly correct.
After generating the four teacher trajectories for each training example, we evaluate the final answer of every trajectory against the corresponding ground-truth answer.
An example is included in the RL training set if \emph{at least one} of its four teacher generations is incorrect.

Specifically, the Stage-II dataset is defined as
\begin{equation}
\mathcal{D}_{\mathrm{RL}}
=
\left\{
x_i \in \mathcal{D}_{\mathrm{train}}
\;\middle|\;
\sum_{j=1}^{4}
\mathbb{I}
\left[
\operatorname{Eval}\!\left(
\hat{a}_{i,j}^{T},
a_i^*
\right)=1
\right]
< 4
\right\}.
\end{equation}

Equivalently, a training example is retained whenever there exists at least one teacher generation that fails to reach the correct final answer.
Examples for which all four teacher generations are correct are excluded from Stage-II training.

This construction concentrates reinforcement learning on examples that remain non-trivial under the teacher-generation process, while reusing the same trajectory-generation results without requiring additional teacher inference.
Note that the teacher trajectories themselves are not used as reward targets during Stage II; once the RL subset is selected, policy optimization is performed using student-generated rollouts under student-side temporal observations, as described in~\cref{subsec:stage_II}.

\section{Experimental Details}
\label{sec:append_experiment}

\subsection{Models and Compared Systems}

We conduct experiments on two Video-LLM families:
Qwen3-VL-4B-Thinking,
Qwen3-VL-8B-Thinking,
InternVL3.5-4B-HF,
and InternVL3.5-8B-HF.
Qwen3-VL is used with its native thinking behavior enabled, whereas InternVL3.5 is used in its standard instruction configuration without enabling an additional explicit thinking mode.
This provides two complementary student reasoning regimes for evaluating reasoning distillation.

For each backbone, we evaluate the pretrained model, Stage-I SFT, standard GRPO, multi-observation GRPO (MO-GRPO), and our full model with Execution-Aware Reference Regularization (EARR).
For the main comparison in~\cref{fig:compare}, we additionally evaluate LUFFY~\citep{yan2026learning} and T-GRPO~\citep{feng2026video} on the 4B students.

Unless otherwise specified, all reinforcement-learning methods are initialized from the same Stage-I SFT checkpoint of the corresponding backbone.
They are trained on the same error-focused RL dataset defined in~\cref{sec:append_data_generate}, and use the same question set, output format, and answer-based reward signal.
This ensures that differences among RL variants arise from their optimization strategies rather than different initialization or training data.

All student models are fine-tuned with LoRA for parameter-efficient adaptation.
We set the LoRA rank to 32, scaling factor to 64, and dropout rate to 0.
LoRA is applied to the
\texttt{q\_proj},
\texttt{k\_proj},
\texttt{v\_proj},
\texttt{o\_proj},
\texttt{gate\_proj},
\texttt{up\_proj},
and \texttt{down\_proj}
modules of the language model.
Training is performed in BF16 precision with SDPA attention.

\subsection{Implementation Details for Training}

\paragraph{Stage-I Supervised Fine-Tuning.}
We train the Stage-I models for 2 epochs with a per-GPU batch size of 1 and 8 gradient accumulation steps.
We use AdamW with a learning rate of $1\times10^{-5}$, zero weight decay, a cosine learning-rate scheduler, and a warmup ratio of 0.03.
The gradient norm is clipped to 1.0, and the maximum sequence length is set to 32,768 tokens.

For Qwen3-VL, videos are processed at 2 FPS with the maximum number of pixels per frame set to 400,704.
For InternVL3.5, we uniformly sample at 2 FPS from each video to control the number of visual tokens.

Stage I optimizes the standard autoregressive cross-entropy objective over the retained correct teacher trajectories, as defined in~\cref{subsec:stage_I}.
The resulting model is used both to initialize Stage II and as the frozen reasoning reference for EARR.

\paragraph{Stage-II Reinforcement Learning.}
All GRPO-based variants are trained for 1 epoch with a learning rate of $1\times10^{-6}$ and zero weight decay.
To control for training resources, all RL-based experiments use the same hardware and wall-clock budget: four NVIDIA H100 GPUs for approximately 48 hours per run, corresponding to approximately 192 H100 GPU-hours.
The RL dataset is described in \cref{sec:append_data_generate}.
We use a per-GPU batch size of 1 and accumulate gradients over four original video-question examples.
The maximum input sequence length and maximum generation length are set to 40,960 and 2,048 tokens, respectively.
The GRPO clipping range is 0.2, and the gradient norm is clipped to 1.0.
Online student rollouts are sampled with temperature 1.0 and top-$p$ 0.95.

All Stage-II methods are trained on the same error-focused RL subset introduced in~\cref{sec:append_data_generate}.
Teacher correctness is used only to determine whether a video-question pair is included in this subset.
Once an example enters Stage II, optimization uses student-generated rollouts and ground-truth task outcomes; teacher trajectories are not used as reward targets.

All RL variants use the same answer- and format-based reward function.
The primary reward is determined by final-answer correctness.
Producing a valid final answer and following the required output format each provide an auxiliary reward of 0.05.
Outputs that reach the maximum generation length without producing a valid answer receive a truncation penalty of 0.5, while excessive repeated 4-grams receive a repetition penalty of 0.2.
No teacher correctness signal or cross-observation difficulty signal is directly included in the rollout reward.

\paragraph{Standard GRPO.}
Standard GRPO uses a single student-side temporal observation for each video-question pair and generates four student rollouts under that observation.
Rewards are normalized within the four-rollout group.
The KL coefficient is fixed to $\beta=0.003$ throughout training.
This configuration serves as the single-observation RL baseline for evaluating the contribution of multi-observation execution.

\paragraph{Multi-Observation Configuration.}
MO-GRPO constructs $K=5$ temporal observations for each video-question pair:
\begin{equation}
\begin{aligned}
&(1\ \mathrm{FPS},0), \qquad
(1\ \mathrm{FPS},0.5), \qquad
&(2\ \mathrm{FPS},0), \qquad
(2\ \mathrm{FPS},0.25), \qquad
(2\ \mathrm{FPS},0.5),
\end{aligned}
\end{equation}
where the second term denotes the sampling phase ratio $\phi$.
For a sampling rate $f$ and phase ratio $\phi$, sampling timestamps are given by $t_k = \frac{k+\phi}{f}.$

Accordingly, the two 1-FPS observations use absolute offsets of 0 and 0.5 seconds,
while the three 2-FPS observations use offsets of 0, 0.125, and 0.25 seconds.
All observations are generated from the same raw video.
Qwen3-VL and InternVL3.5 use the same timestamp-based frame-selection logic so that the observation configurations have consistent temporal semantics across model families.

\paragraph{Observation-Local Optimization.}
For MO-GRPO without EARR, each of the five temporal observations independently generates four student rollouts.
Reward statistics and group-relative advantages are computed only among rollouts generated from the same observation; rewards are never normalized across different temporal observations.

With four original video-question pairs accumulated per optimizer update,
This variant uses a fixed KL coefficient of $\beta=0.003$.

\paragraph{Execution-Aware Reference Regularization.}
For the full model, we preserve the same total generation budget of four generations per observation.
Each observation produces four stochastic student rollouts together with one deterministic response from the frozen Stage-I reference policy.

The reference response uses the same temporal observation, question, and maximum generation length as the student rollouts, but is decoded deterministically with temperature 0 and top-$p$ 1.0.
Its final-answer correctness is used only to determine the observation-specific reference coefficient:
\begin{equation}
\beta_{i,k}=
\begin{cases}
\beta_{\mathrm{high}}=0.005,
& \text{if the reference answer is correct},\\
\beta_{\mathrm{low}}=0.001,
& \text{otherwise}.
\end{cases}
\end{equation}

The reference response is not included in the GRPO rollout group and does not participate in the computation of the reward mean, standard deviation, group-relative advantage, or policy loss.
These quantities are computed exclusively from the four student rollouts generated under the same observation.


\subsection{Baselines}
\label{subsec:appendix_baselines}

We compare against supervised reasoning distillation and representative RL-based reasoning optimization methods.
All RL baselines use the same Stage-I initialization and Stage-II training subset whenever their original formulation permits, and we keep the student backbone, training questions, answer format, and evaluation protocol fixed.

\paragraph{SFT.}
SFT~\citep{ouyang2022training} directly trains the student on correct teacher-generated reasoning trajectories using the Stage-I supervised objective described above.
The same LoRA configuration is used as in our method.

\paragraph{GRPO.}
GRPO~\citep{guo2025deepseek} performs on-policy reinforcement learning with group-relative reward normalization over multiple student rollouts.
In our implementation, it operates on a single temporal observation per video-question pair and uses the same Stage-I SFT initialization and Stage-II RL dataset as our method.

\paragraph{SFT $\rightarrow$ GRPO.}
This baseline follows the same two-stage initialization used by our framework but applies standard single-observation GRPO in Stage II.
Importantly, it uses the same Stage-I SFT checkpoint and the same error-focused RL training set as our method; no separate data split is introduced between the SFT and RL stages.

\paragraph{LUFFY.}
LUFFY~\citep{yan2026learning} is an off-policy reasoning optimization method that combines external teacher trajectories with on-policy student rollouts.
We adapt LUFFY to the same video reasoning training pool and use the same teacher trajectories available to the other distillation methods.
The student initialization, task data, output format, and evaluation protocol are kept consistent with the other RL baselines.

\paragraph{T-GRPO.}
T-GRPO~\citep{feng2026video} introduces temporal awareness into GRPO by contrasting model behavior under temporally ordered and shuffled video frames.
We implement its temporal reward on the same Stage-II examples and student backbone used by the other RL methods while keeping the remaining training and evaluation settings unchanged.

\subsection{Evaluation Protocol}
\label{subsec:append_eval}
All pretrained and post-trained models are evaluated using the same unified evaluation pipeline.
At inference time, the model receives only the raw video and benchmark question; ground-truth answers are used solely for post-generation evaluation.
Both Qwen3-VL and InternVL3.5 are evaluated using a fixed 2-FPS temporal sampling configuration with deterministic decoding:
temperature 0, top-$p$ 1.0, and a maximum of 2,048 newly generated tokens.

For TimeBlind~\citep{li2026timeblind},
MVBench~\citep{li2024mvbench},
TempCompass~\citep{liu2024tempcompass},
TOMATO~\citep{shangguan2025tomato},
and the Short subset of Video-MME~\citep{fu2025video},
we extract the final answer from the required \texttt{Answer:} field and evaluate it against the benchmark ground truth using the corresponding deterministic answer-normalization rule.

VideoReasonBench~\citep{liu2026videoreasonbench} contains open-ended reasoning and structured state-prediction questions.
We follow a fixed evaluation pipeline for all candidate models and use the same judge configuration and prompts throughout evaluation.
For structured state-prediction questions, the generated response is first converted into the required structured representation before the benchmark's state-transition verification is applied.
No candidate model receives access to the evaluation reference answer during generation.

\subsection{Observation-Sensitivity Analysis}
\label{subsec:append_observation_sensitivity_details}

The analysis in \cref{fig:observation_sensitivity} is an inference-only
diagnostic experiment on the held-out TimeBlind~\citep{li2026timeblind} evaluation set and does
not involve any additional training or post-training.
We evaluate Qwen3-VL-4B, Qwen3-VL-8B~\citep{bai2025qwen3}, InternVL3.5-4B, InternVL3.5-8B~\citep{wang2025internvl3}, and InternVideo3-8B~\citep{yan2026internvideo3agentifyfoundationmodels} on the same 800 held-out video-question pairs.
Within each model, the raw video, question, prompt, answer extraction, spatial preprocessing, and decoding configuration are kept fixed; only the temporal sampling configuration is changed.
All observations are evaluated independently using deterministic decoding with temperature $0$ and top-$p=1$.
We use the same TimeBlind prompt and deterministic answer evaluation described in \cref{subsec:append_eval,subsec:append_prompt}.

\paragraph{Observation-dependent execution.}
For \cref{fig:observation_sensitivity}(a), we construct five temporal
observations for every video-question pair:
\[
(1,0),\ (1,0.5),\ (2,0),\ (2,0.25),\ (2,0.5),
\]
where each tuple denotes $(f,\rho)$, with sampling rate $f$ in FPS and
sampling phase $\rho$.
For a fixed model and question, we record binary correctness under each
of the five observations.
A question is categorized as \emph{always incorrect} if all five
predictions are incorrect, \emph{always correct} if all five are
correct, and \emph{observation-dependent} otherwise.
We report the fraction of held-out TimeBlind questions in each category.

\paragraph{Outcome flips under phase shifts.}
For \cref{fig:observation_sensitivity}(b), we vary the sampling rate over $f\in\{0.5,1,2,4\}$ FPS while comparing two sampling phases, $\rho\in\{0,0.5\}$.
For a given $(f,\rho)$, sampling timestamps follow$t_k = \frac{k+\rho}{f}.$
At each fixed sampling rate, an outcome flip occurs when the same model changes from correct to incorrect, or vice versa, between the two phase settings.
For model $m$, the flip rate at sampling rate $f$ is

\begin{equation}
    \mathrm{FlipRate}_m(f)=\frac{1}{N}\sum_{i=1}^{N}\mathbb{I}\left[c_{m,i}^{(f,0)}\neq c_{m,i}^{(f,0.5)}\right],
\end{equation}

where $N=800$ and $c_{m,i}^{(f,\rho)}\in\{0,1\}$ denotes answercorrectness.
The \emph{Average} curve in \cref{fig:observation_sensitivity}(b) reports the macro-average of the flip rates over the five evaluated models.

\subsection{Prompt}
\label{subsec:append_prompt}

We provide the complete benchmark-specific input prompts used in our experiments.
The same task prompt and output format are used during knowledge distillation and evaluation for each benchmark.

We do not explicitly request a thinking trace in the user prompt.
Instead, the user prompt specifies only the task, the evidence field, and the final-answer field.
For models with a native reasoning mechanism, such as Qwen3-VL-Thinking, intermediate reasoning is produced through the model's native thinking behavior.
InternVL3.5 is used in its standard instruction configuration.

\begin{figure*}[ht!]
    \centering
    \includegraphics[width=0.9\textwidth]{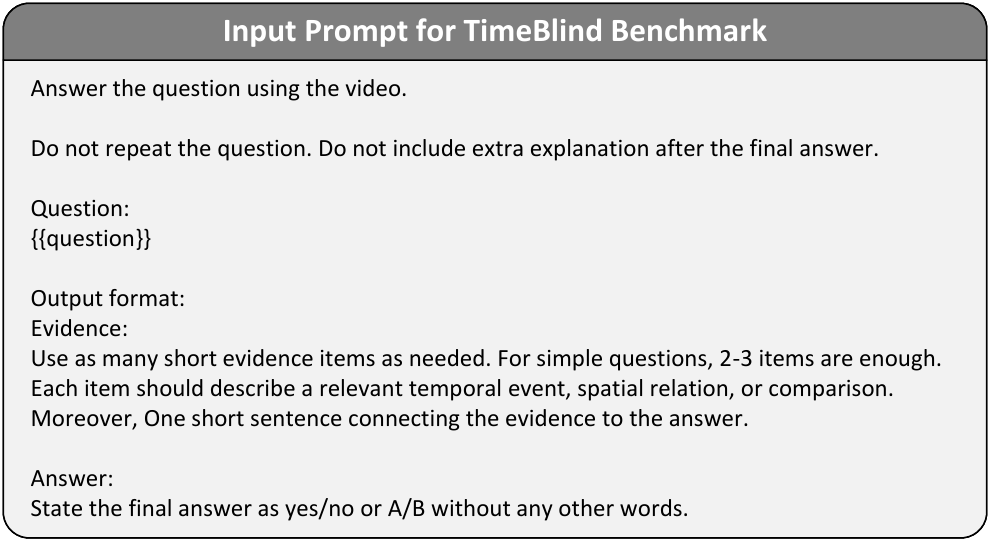}
    \caption{Prompt Template for TimeBlind Benchmark.}
    \label{fig:prompt_timeblind}
\end{figure*}

\begin{figure*}[ht!]
    \centering
    \includegraphics[width=0.9\textwidth]{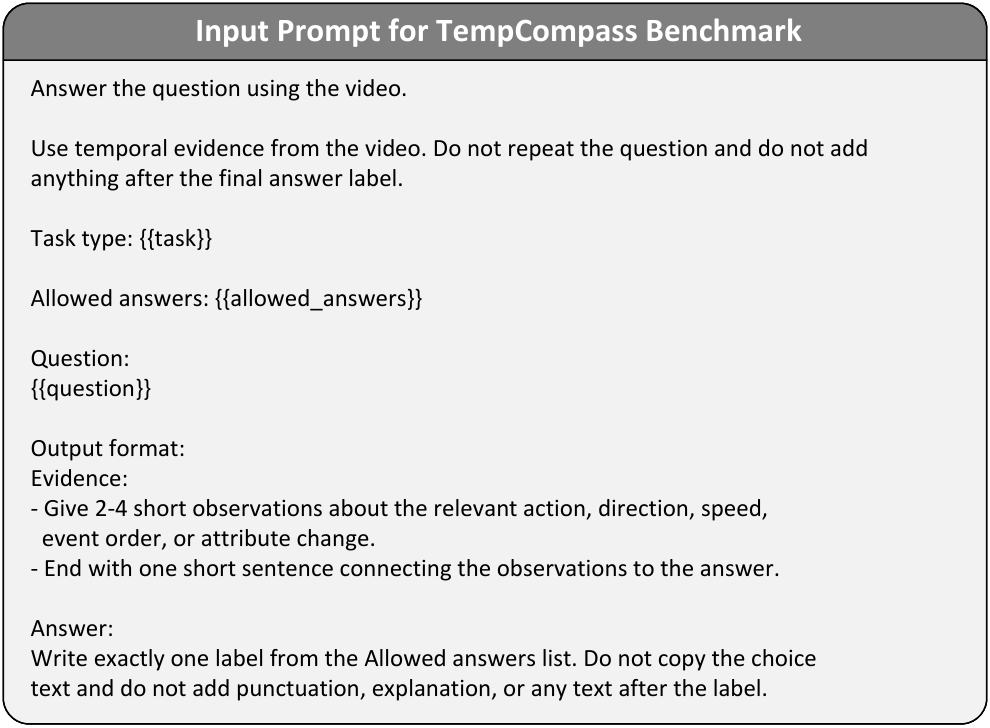}
    \caption{Prompt Template for TempCompass Benchmark.}
    \label{fig:prompt_TempCompass}
\end{figure*}

\begin{figure*}[ht!]
    \centering
    \includegraphics[width=0.9\textwidth]{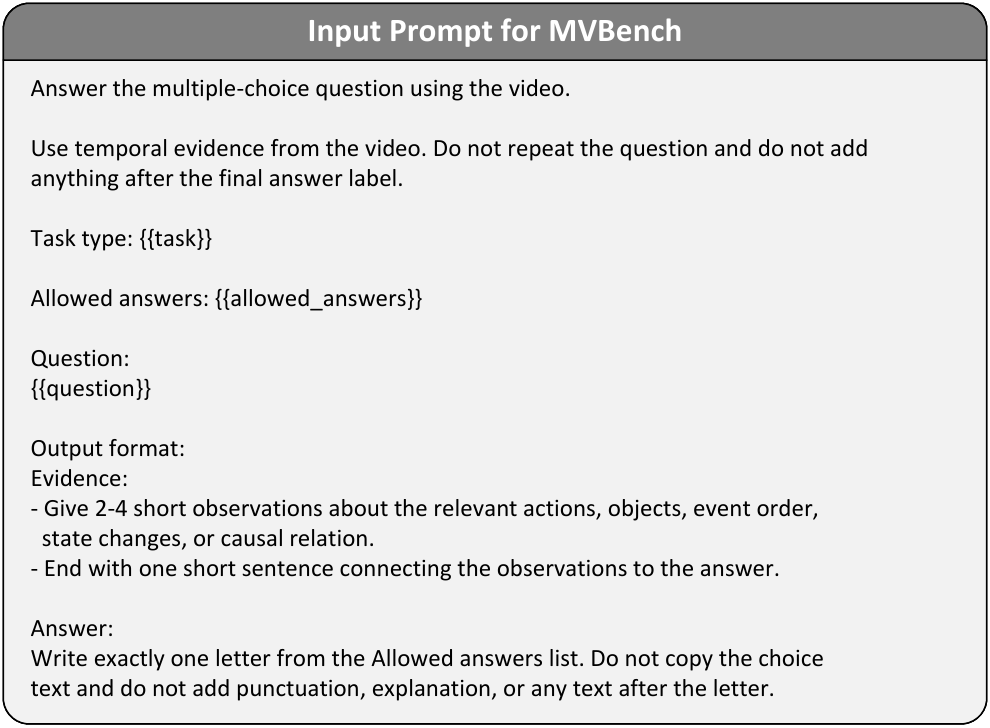}
    \caption{Prompt Template for MVBench Benchmark.}
    \label{fig:prompt_MVBench}
\end{figure*}

\begin{figure*}[ht!]
    \centering
    \includegraphics[width=0.9\textwidth]{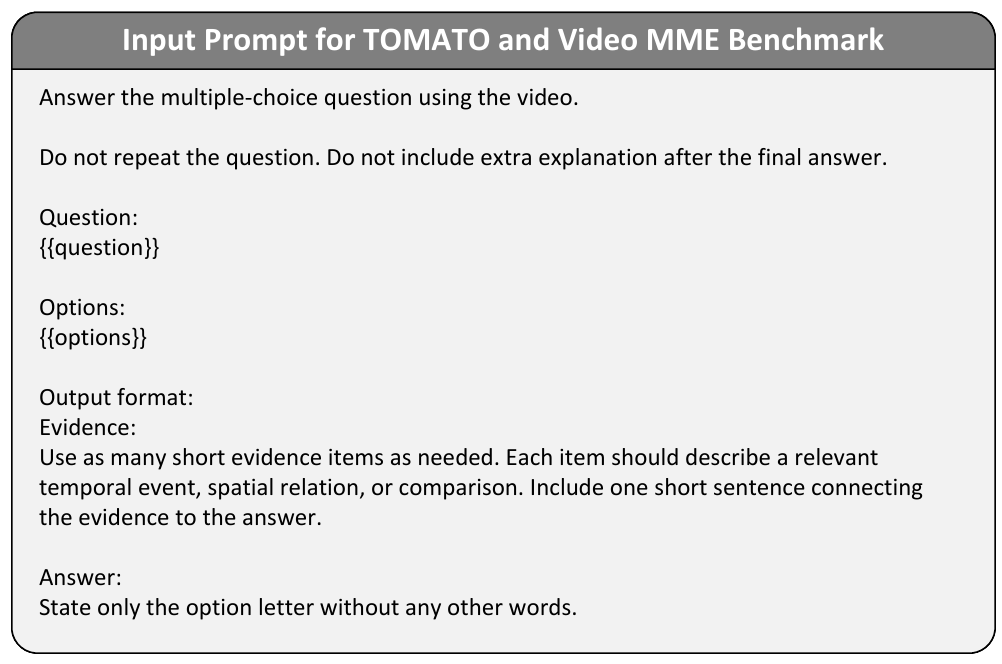}
    \caption{Prompt Template for TOMATO and Video-MME Benchmark.}
    \label{fig:prompt_ood_bench}
\end{figure*}

\begin{figure*}[ht!]
    \centering
    \includegraphics[width=0.9\textwidth]{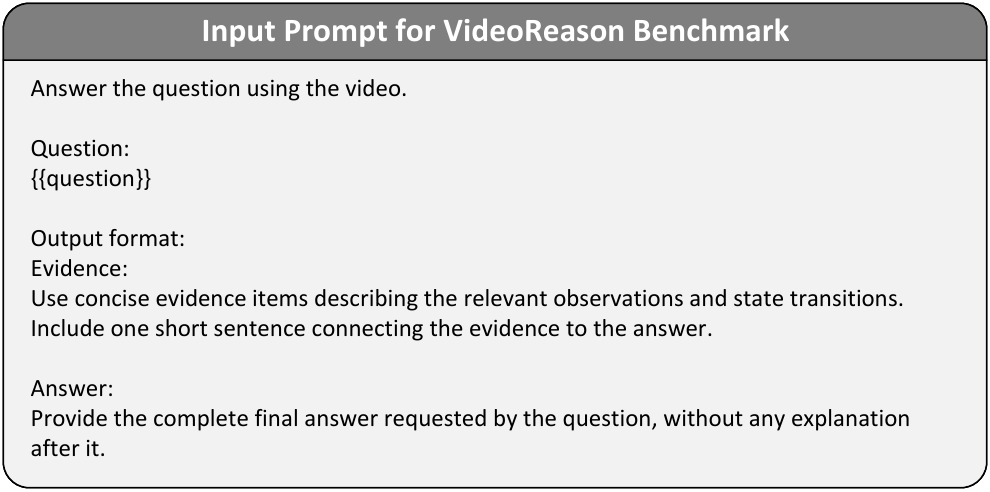}
    \caption{Prompt Template for VideoReason Benchmark.}
    \label{fig:prompt_VideoReason}
\end{figure*}

TimeBlind uses the prompt in~\cref{fig:prompt_timeblind},
TempCompass uses~\cref{fig:prompt_TempCompass},
MVBench uses~\cref{fig:prompt_MVBench},
TOMATO and Video-MME (Short) share the prompt in~\cref{fig:prompt_ood_bench},
and VideoReasonBench uses~\cref{fig:prompt_VideoReason}.

\begin{figure*}[ht!]
    \centering
    \includegraphics[width=0.9\textwidth]{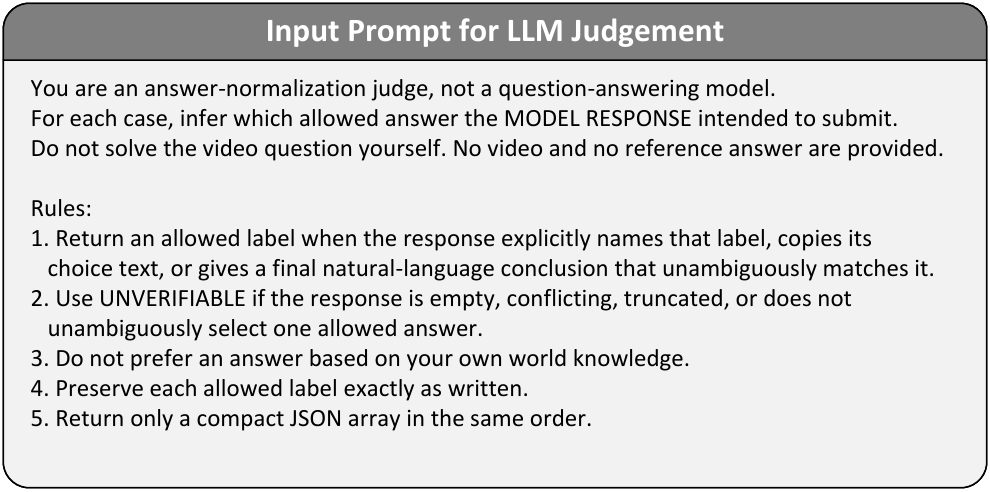}
    \caption{Prompt Template for LLM Judgment.}
    \label{fig:prompt_LLM_Judge}
\end{figure*}

The answer-normalization prompt shown in~\cref{fig:prompt_LLM_Judge} is used only to map ambiguous free-form model outputs to the benchmark's allowed answer labels.
It is not provided with the video or reference answer and is not used to solve the underlying video question.

\begin{figure*}[t!]
    \vspace{-4mm}
    \centering
    \includegraphics[height=0.95\textheight,keepaspectratio]{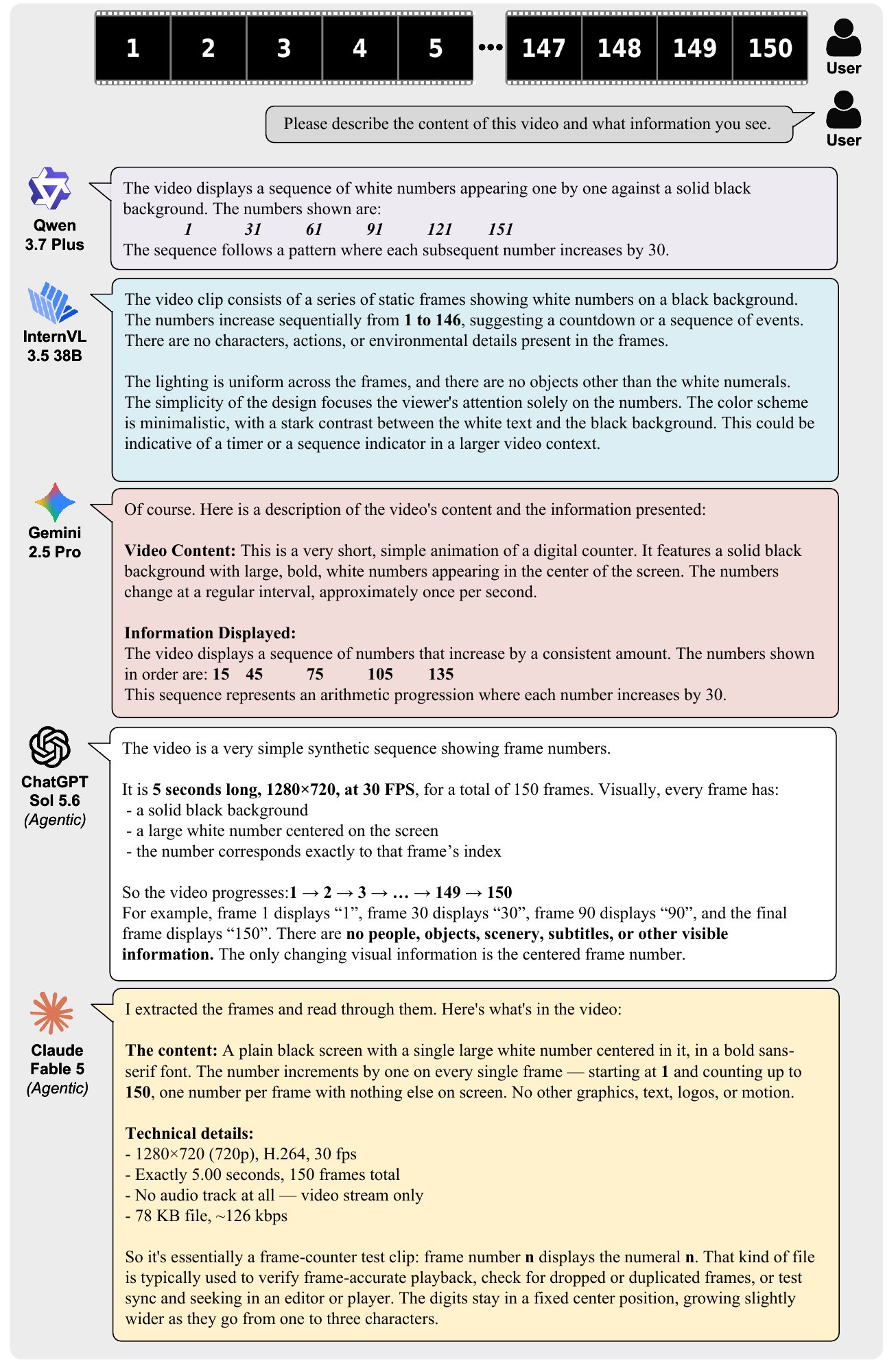}
    \caption{\textbf{Case study of frame-level temporal perception.} Standard Video-LLMs misinterpret sparsely sampled frames, while agentic models recover the full frame sequence.}
    \label{fig:case_study_2}
\end{figure*}

\begin{figure*}[t!]
    \vspace{-4mm}
    \centering
    \includegraphics[height=0.95\textheight,keepaspectratio]{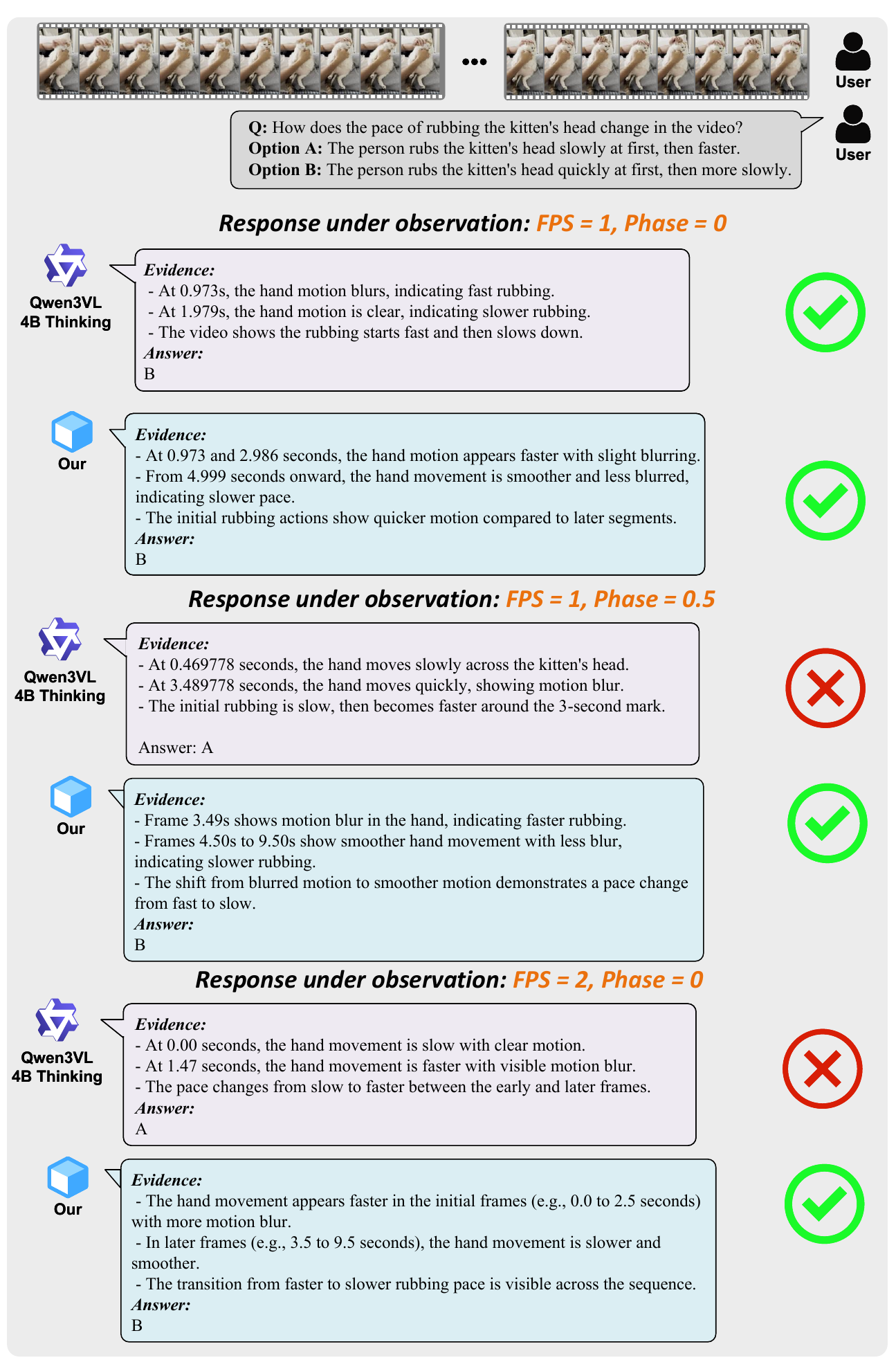}
    \caption{\textbf{Case Study 1 (Part I)} Comparison between Qwen3-VL-4B-Thinking and our method under different temporal observations of the same video-question pair.}
    \label{fig:case_study_1_1}
\end{figure*}

\begin{figure*}[t!]
    \vspace{-4mm}
    \centering
    \includegraphics[width=\textwidth]{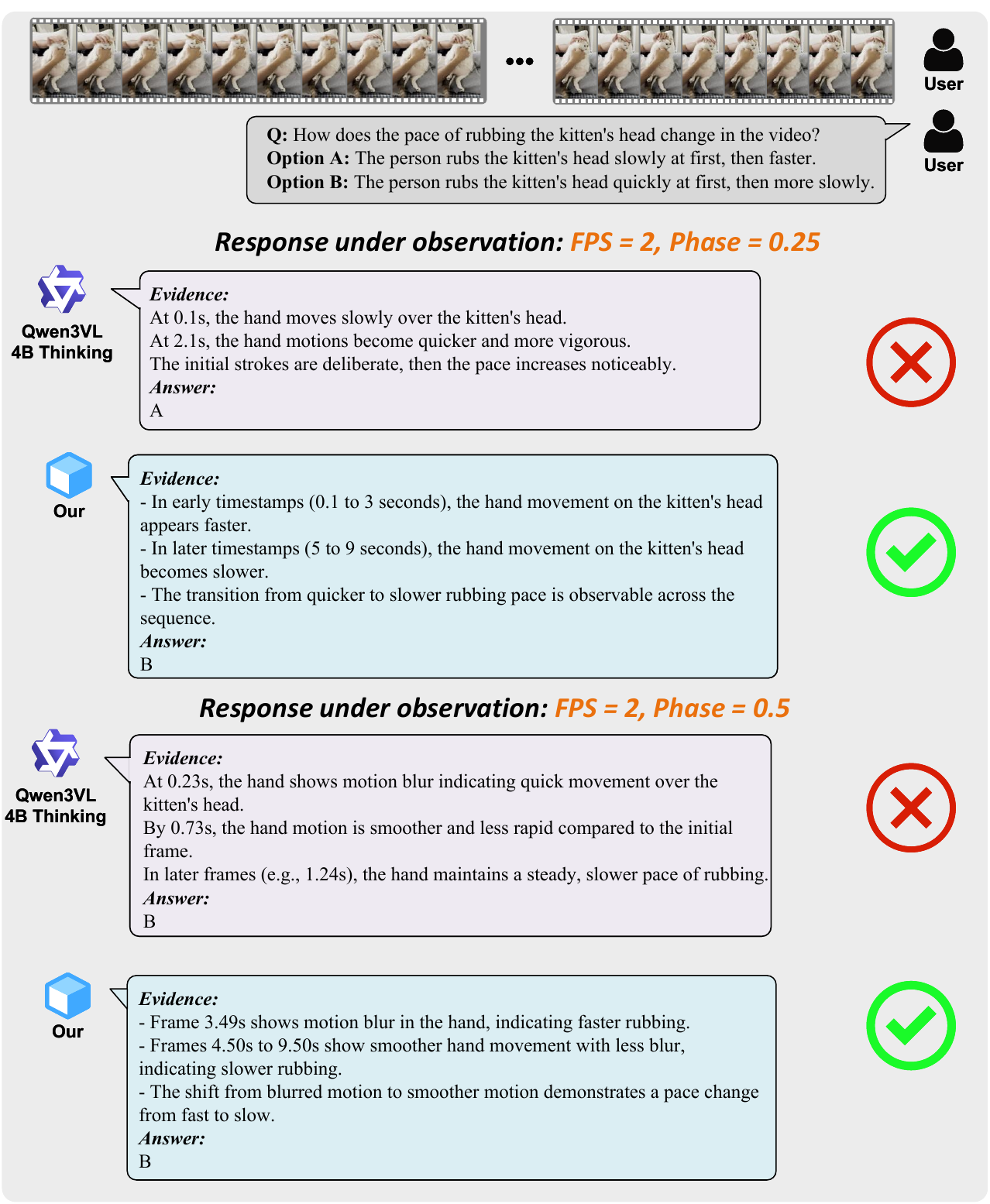}
    \caption{\textbf{Case Study 1 (Part II)} Additional comparison between Qwen3-VL-4B-Thinking and our method under different temporal observations of the same video-question pair.}
    \label{fig:case_study_1_2}
\end{figure*}

\section{Additional Studies}
\label{sec:append_case_study}

\subsection{Reasoning under Different Temporal Observations}
To qualitatively complement the observation-sensitivity analysis in \cref{sec:problem}, we examine a TimeBlind example that requires comparing the rubbing speed of a hand over time. 
\cref{fig:case_study_1_1} and~\cref{fig:case_study_1_2} show responses from Qwen3-VL-4B-Thinking~\citep{bai2025qwen3} and our model under different temporal observations obtained by varying the sampling FPS and phase. 
The underlying video and question remain unchanged; only the temporal evidence exposed to the model differs.

The baseline exhibits observation-dependent execution: its prediction changes across sampling configurations, and under several observations it reverses the temporal trend from fast-to-slow'' to slow-to-fast.'' In contrast, our model consistently predicts the correct transition while grounding its reasoning in the evidence available under each observation.
Importantly, the evidence itself is not identical across observations: the model refers to different timestamps and motion cues depending on the sampled frames. 
This illustrates the intended behavior of observation-conditioned execution—rather than reproducing a fixed reasoning trajectory, the model adapts its reasoning path to the current temporal evidence.

\subsection{Observation Mismatch in Practice}

\cref{fig:case_study_2} provides a controlled illustration of the observation mismatch discussed in~\cref{sec:problem}.
We construct a five-second synthetic video containing 150 consecutive frames, where frame $t$ displays the integer $t$.
Because the underlying visual sequence is deterministic and contains no semantic ambiguity, differences in model responses can be directly related to the temporal evidence exposed to each model.

As shown in~\cref{fig:case_study_2}, standard Video-LLMs do not necessarily recover the complete temporal sequence from the same raw video.
For example, different models report sparse or shifted subsequences rather than the full progression from 1 to 150.
These responses reflect the frames made available by their internal video sampling procedures: changing the sampling rate, phase, or frame locations produces a different effective observation even though the underlying video remains unchanged.
In contrast, the agentic systems shown in the figure explicitly inspect the video frames and recover the complete sequence.
We emphasize that this comparison is not intended as a direct capability ranking between standard and agentic models, since the latter have access to additional frame-inspection tools.
Rather, the example isolates the central issue considered in this work: the raw video $V$ is shared, while the effective model observation
$O=S(V;\xi)$ can differ substantially across systems.

This distinction becomes particularly important for reasoning distillation.
A teacher trajectory is generated from the evidence contained in its observation $O_T$, while the student is trained to reproduce that trajectory from a potentially different observation $O_S$.
If the teacher reasoning explicitly relies on an event or frame that is absent from $O_S$, the trajectory can remain logically correct under $O_T$ while being difficult or impossible to execute from the evidence available to the student.
The issue is therefore not necessarily an incorrect teacher rationale, but a mismatch between the observation under which the rationale was generated and the observation under which it is subsequently learned.
This is the conditional target mismatch introduced in~\cref{sec:problem}.

\begin{table*}[t!]
\centering

\caption{\textbf{Results with an open-source teacher model.}
Distillation performance using Qwen3-VL-235B-A22B-Thinking as the teacher and InternVL3.5-4B and Qwen3-VL-Thinking-4B as student models across three in-domain and three out-of-domain benchmarks.
Consistent improvements over the corresponding pretrained and SFT baselines indicate that the proposed framework remains effective with an open-source teacher, supporting the reproducibility of the main findings.}


\small
\setlength{\tabcolsep}{4.3pt}
\renewcommand{\arraystretch}{1.1}
\resizebox{\textwidth}{!}{
\begin{tabular}{ccccccc}
\hline
                         & \multicolumn{3}{c}{\textit{\textbf{In-Domain}}}  & \multicolumn{3}{c}{\textit{\textbf{Out-of-Domain}}} \\ \cline{2-7} 
\multirow{-2}{*}{Method} & TimeBlind      & MVBench        & TempCompass    & TOMATO           & Video-MME       & VideoReason    \\ \hline
\multicolumn{7}{c}{\textit{\textbf{Teacher Model: Qwen3-VL-235B-A22B-Thinking}}}                                                   \\ \hline
-                        & 69.75          & 71.38          & 84.77          & 44.85            & 88.50           & 27.48          \\ \hline
\multicolumn{7}{c}{\textit{\textbf{Student Model: InternVL 3.5 4B}}}                                                              \\ \hline
Baseline                 & 47.88          & 57.00          & 67.84          & 21.29            & 67.00           & 0.97           \\
CoT SFT                  & 50.14          & 58.98          & 65.34          & 23.70            & 68.24           & 1.12           \\
\rowcolor[HTML]{EFEFEF} 
\textbf{Ours}            & \textbf{60.13} & \textbf{59.50} & \textbf{77.67} & \textbf{30.80}   & \textbf{70.44}  & \textbf{2.76}  \\ \hline
\multicolumn{7}{c}{\textit{\textbf{Student Model: Qwen3-VL-Thinking 4B}}}                                                         \\ \hline
Baseline                 & 64.50          & 66.63          & 79.69          & 32.01            & 76.67           & 5.21           \\
SFT                      & 63.45          & 65.42          & 77.92          & 24.30            & 74.26           & 4.88           \\
\rowcolor[HTML]{EFEFEF} 
\textbf{Ours}            & \textbf{66.75} & \textbf{67.20} & \textbf{80.96} & \textbf{34.97}   & \textbf{78.91}  & \textbf{7.82}  \\ \hline
\end{tabular}}
\label{tab:open_source_result}

\vspace{-2mm}
\end{table*}

\subsection{Reproducibility with an Open-Source Teacher}
\label{subsec:opensource_teacher}

To further examine whether the main findings depend on a specific proprietary teacher, we replace Gemini 2.5 Pro with the open-source Qwen3-VL-235B-A22B-Thinking model~\citep{bai2025qwen3}.
Teacher inference is performed through the OpenRouter API~\citep{openrouter2023}, with videos processed at 2 FPS to generate reasoning trajectories for distillation.
The resulting trajectories are then used to train InternVL3.5-4B~\citep{wang2025internvl3} and Qwen3-VL-4B-Thinking~\citep{bai2025qwen3} following the same acquisition-to-execution pipeline as in the main experiments.
All remaining dataset splits, training procedures, and evaluation protocols are kept unchanged, with performance evaluated on the same three in-domain and three out-of-domain video reasoning benchmarks.

As shown in \cref{tab:open_source_result}, the overall behavior observed in the main experiments persists after replacing the proprietary teacher with an open-source model.
Direct trajectory SFT remains strongly dependent on the student reasoning regime: teacher trajectories provide useful supervision for InternVL3.5 on several tasks, while direct imitation can still degrade the pretrained Qwen3-VL-Thinking model.
In contrast, the proposed acquisition-to-execution framework consistently outperforms both the corresponding pretrained and SFT models across all evaluated in-domain and out-of-domain settings.
These results indicate that the benefit of observation-conditioned execution is not specific to Gemini-generated trajectories, while providing an open-source teacher setting for reproducing the main experimental findings.

\end{document}